\documentclass{article}

\usepackage{microtype}
\usepackage{graphicx}
\usepackage{subfigure}
\usepackage{booktabs}
\usepackage{array}
\usepackage{amsmath}
\usepackage{xcolor}
\usepackage{xspace}
\usepackage{hyperref}
\pdftrailerid{}
\usepackage{tikz}
\usepackage{pgfplots}
\pgfplotsset{compat=1.16}

\usepackage[accepted]{mlsys2025}
\makeatletter\renewcommand{\Notice@String}{Preprint.}\makeatother
\hypersetup{pdfsubject={}}

\newcommand{\system}{\textsc{Cap-and-Track}\xspace}
\newcommand{\foqs}{\textsc{FOQS}\xspace}
\newcommand{\ride}{\textsc{Track}\xspace}
\newcommand{\bound}{\textsc{Cap}\xspace}
\newcommand{\rqFixSrc}{} %
\newcommand{\rqRideFullG}{108.5\xspace} %
\newcommand{\rqRideFullShare}{3.8\%\xspace} %
\newcommand{\rqHighFullG}{109.5\xspace} %
\newcommand{\rqHighFullShare}{15.5\%\xspace} %
\newcommand{\rqHighCutShare}{58\%\xspace} %
\newcommand{\rqFixFullG}{107.0\xspace} %
\newcommand{\rqFixFullShare}{1.1\%\xspace} %
\newcommand{\rqFixCutShare}{49\%\xspace} %
\newcommand{\rqCapGap}{about 0.8\%\xspace} %
\newcommand{\rqTraceRuns}{Median of three runs per arm; bands span the runs.\xspace} %
\newcommand{\rqAbsGap}{1\%\xspace} %
\newcommand{\rqRejRatio}{about a quarter\xspace} %
\newcommand{\apBacklog}{\ref{app:backlog}\xspace}
\newcommand{\apBatch}{\ref{app:batch}\xspace}
\newcommand{\apScale}{\ref{app:scale}\xspace}
\newcommand{\apFleetsrc}{\ref{app:fleetsrc}\xspace}
\newcommand{\apCoord}{\ref{app:coordination}\xspace}
\newcommand{\apFigAimd}{\ref{fig:aimd}\xspace}
\newcommand{\apRevTab}{\ref{tab:revfixed}\xspace}
\newcommand{\apRefusal}{\ref{app:refusal}\xspace}
\newcommand{\apReserved}{\ref{app:reserved}\xspace}
\newcommand{\apDialup}{\ref{app:dialup}\xspace}
\newcommand{\apBounds}{\ref{app:bounds}\xspace}
\newcommand{\apEpisodes}{\ref{app:episodes}\xspace}
\newcommand{\apReplay}{\ref{app:replay}\xspace}
\newcommand{\apCodel}{\ref{app:codel}\xspace}
\newcommand{\apBudget}{\ref{app:budget}\xspace}
\newcommand{\apCtrl}{\ref{app:ctrl}\xspace}
\newcommand{\apRevfixed}{\ref{app:revfixed}\xspace}
\newcommand{\apRamp}{\ref{app:ramp}\xspace}
\newcommand{\apPercap}{\ref{app:percap}\xspace}
\newcommand{\apAutoscale}{\ref{app:autoscale}\xspace}
\newcommand{\apFresh}{\ref{app:fresh}\xspace}
\newcommand{\apSafe}{\ref{app:safe}\xspace}

\mlsystitlerunning{Cap-and-Track: LLM Inference on Shifting Capacity}

\newcommand{\rqTARideFullG}{108.5\xspace} %
\newcommand{\rqTARideFullR}{0.04\xspace} %
\newcommand{\rqTARideCutG}{54.7\xspace} %
\newcommand{\rqTARideCutR}{0.49\xspace} %
\newcommand{\rqTARideRestG}{102.3\xspace} %
\newcommand{\rqTARideRestR}{0.03\xspace} %
\newcommand{\rqTARideRec}{220\xspace} %
\newcommand{\rqTARideQue}{0.54\xspace} %
\newcommand{\rqTAFixFullG}{107.0\xspace} %
\newcommand{\rqTAFixFullR}{0.01\xspace} %
\newcommand{\rqTAFixCutG}{54.9\xspace} %
\newcommand{\rqTAFixCutR}{0.97\xspace} %
\newcommand{\rqTAFixRestG}{107.1\xspace} %
\newcommand{\rqTAFixRestR}{0.01\xspace} %
\newcommand{\rqTAFixRec}{0\xspace} %
\newcommand{\rqTAFixQue}{0.34\xspace} %
\newcommand{\rqTAHighFullG}{109.5\xspace} %
\newcommand{\rqTAHighFullR}{0.18\xspace} %
\newcommand{\rqTAHighCutG}{54.9\xspace} %
\newcommand{\rqTAHighCutR}{1.35\xspace} %
\newcommand{\rqTAHighRestG}{109.8\xspace} %
\newcommand{\rqTAHighRestR}{0.18\xspace} %
\newcommand{\rqTAHighRec}{0\xspace} %
\newcommand{\rqTAHighQue}{0.65\xspace} %
\newcommand{\rqTALowFullG}{54.5\xspace} %
\newcommand{\rqTALowFullR}{0.00\xspace} %
\newcommand{\rqTALowCutG}{54.0\xspace} %
\newcommand{\rqTALowCutR}{0.01\xspace} %
\newcommand{\rqTALowRestG}{54.5\xspace} %
\newcommand{\rqTALowRestR}{0.00\xspace} %
\newcommand{\rqTALowRec}{--\xspace} %
\newcommand{\rqTALowQue}{0.07\xspace} %
\newcommand{\rqTBRideFullG}{108.9\xspace} %
\newcommand{\rqTBRideFullR}{0.05\xspace} %
\newcommand{\rqTBRideCutG}{55.2\xspace} %
\newcommand{\rqTBRideCutR}{0.48\xspace} %
\newcommand{\rqTBRideRestG}{102.7\xspace} %
\newcommand{\rqTBRideRestR}{0.04\xspace} %
\newcommand{\rqTBRideRec}{209\xspace} %
\newcommand{\rqTBRideQue}{0.92\xspace} %
\newcommand{\rqTBFixFullG}{107.1\xspace} %
\newcommand{\rqTBFixFullR}{0.01\xspace} %
\newcommand{\rqTBFixCutG}{54.9\xspace} %
\newcommand{\rqTBFixCutR}{0.97\xspace} %
\newcommand{\rqTBFixRestG}{107.2\xspace} %
\newcommand{\rqTBFixRestR}{0.01\xspace} %
\newcommand{\rqTBFixRec}{0\xspace} %
\newcommand{\rqTBFixQue}{0.37\xspace} %
\newcommand{\rqTBHighFullG}{109.2\xspace} %
\newcommand{\rqTBHighFullR}{0.19\xspace} %
\newcommand{\rqTBHighCutG}{54.7\xspace} %
\newcommand{\rqTBHighCutR}{1.36\xspace} %
\newcommand{\rqTBHighRestG}{109.5\xspace} %
\newcommand{\rqTBHighRestR}{0.18\xspace} %
\newcommand{\rqTBHighRec}{0\xspace} %
\newcommand{\rqTBHighQue}{1.07\xspace} %
\newcommand{\rqRuns}{three\xspace} %

\begin{document}

\twocolumn[
\mlsystitle{Cap-and-Track: Optimizing LLM Inference Request Dispatch Across Shifting Capacity}

\mlsyssetsymbol{now}{\dag}
\begin{mlsysauthorlist}
\mlsysauthor{Benjamin Novotny}{meta}
\mlsysauthor{Doug Szeto}{meta}
\mlsysauthor{Chang Wang}{meta}
\mlsysauthor{Shicheng Chen}{meta,now}
\mlsysauthor{George Zhihua Yuan}{meta}
\mlsysauthor{Bob Kamma}{meta}
\mlsysauthor{Michael Ge}{meta}
\mlsysauthor{John Xin Zhou}{meta}
\mlsysauthor{Andrea Olgiati}{meta}
\end{mlsysauthorlist}

\mlsysaffiliation{meta}{Meta}
\mlsyscorrespondingauthor{Benjamin Novotny}{benovo@meta.com}
\mlsyskeywords{LLM serving, admission control, overload control, durable queues}

\vskip 0.3in

\begin{abstract}
A fixed dispatch rate for LLM inference becomes stale as serving fleets resize and request sizes change. Generative output length is unknown at admission, and shared provider pools can lose capacity. We present \system, a design for asynchronous inference across provider capacity and internal GPUs. Meta's traffic shaper implements \system for post and live-video understanding. Refusals signal the backend's current admission ceiling. Durable priority queues hold excess work outside execution, including refused requests for retry. \bound is a fail-fast admission check at the saturated resource that refuses work before a long wait. Using only those refusals, the \ride dispatcher probes upward and backs off. It follows the ceiling without output-length prediction or a fleet-level rate to retune. In a concurrent same-traffic production A/B, the shaper ran per-replica \bound plus \ride. It served a classifier on 43\% fewer H100s than its queued, fixed-rate precursor, though lower-priority work waited longer at backfill peaks. In steady-capacity real-engine replay, \ride ran within \rqAbsGap of the goodput of a fixed rate tuned above measured capacity, with \rqRejRatio of that rate's refusal share. The shaper carries more than 300 million online requests and hundreds of terabytes of media a day. During a backfill, it handled 1.1 billion requests in its busiest 24 hours. Lessons include keeping queues inside inference engines just long enough to fill batches, steering dispatch by capacity refusals alone, keeping dispatch stateful, and sizing fleets by how long low-priority work can wait.
\end{abstract}
]

\printAffiliationsAndNotice{\textsuperscript{\dag}Now at Google (scchen@google.com). }

\section{Introduction}
\label{sec:intro}

The traffic shaper is a shared internal dispatch service for online, asynchronous LLM inference in Meta's social media applications, including post and per-minute live-video understanding. We study four of these workloads in depth, plus two that joined later (Section~\ref{sec:scale}). Each day the shaper carries more than 300 million online requests and stages hundreds of terabytes of media. As load shifts, the shaper decides how much work each inference engine gets. The wrong amount leaves engines idle or overloaded. Accelerators and provider capacity are scarce and slow to add (Section~\ref{sec:setting}), so both should stay busy. For an engine we run, busy means a full batch. Work waiting beyond what keeps that batch full adds only delay and timeouts. When the shaper and its hand-tuned precursor ran at their dispatch limits, the precursor kept its engines' batches only about a third full (Section~\ref{sec:fleet}). Staged services, production overload controllers, and receiver-issued credits have long bounded waiting work~\cite{Welsh2001SEDA,Zhou2018DAGOR,Cho2020Breakwater}.

In LLM serving, demand and supply shift. Demand, the work callers send, swings in volume with daily cycles, events, and backfills (bulk jobs over existing content, launched at any time). Request size depends on an output length unknown a priori~\cite{Kwon2023PagedAttention} and shifts with the token mix and reasoning length. Supply, the serving capacity behind a backend, moves as shared pools congest, replicas are preempted, and fleets resize. Continuous batching rewards concurrency~\cite{Yu2022Orca}. But past the throughput plateau, more concurrency only adds queueing delay and KV-cache pressure~\cite{Kwon2023PagedAttention}. Prefill and decode stress different resources~\cite{Zhong2024DistServe,Agrawal2024SarathiServe}. We measure capacity by goodput, the rate of requests completed within their timeout.

Production stacks scatter admission across tiers. A provider's admission refuses work. A gateway accepts work whenever it can take a connection. An engine queues sequences. Our design rule keeps unbounded waiting out of execution, so neither networks nor engines hold more work than they can serve. Each engine keeps a short queue behind its running batch. Everything else waits in \foqs, Meta's durable priority queue. \bound, a lightweight, fail-fast admission check at the resource that saturates, surfaces overload as a refusal (HTTP 429), so the dispatcher need not predict request size. Providers already return such refusals. For internal engines, we enabled the model server's per-replica cap, so total admission scales with the fleet (Section~\ref{sec:design}). vLLM 0.29, released after we deployed, adds a queue limit like that cap, off by default~\cite{vLLM029}. A limit only refuses, so a short engine queue needs a dispatcher that follows the limit closely. A looser dispatcher needs a deeper queue, which exposes more in-flight work to timeouts and dropped connections (Section~\ref{sec:lessons}). \ride is the close-following dispatcher, a dedicated dispatch loop per capacity domain (a provider quota domain or serving pool). \ride learns the ceiling, the highest rate the backend currently serves, only from the \bound's refusals. The durable queue drains only at \ride's rate.

The shaper implements \system. The four main workloads, named by their capacity or model, all serve post and live-video understanding. The reserved and on-demand workloads run on matching provider capacity. The classifier and zero-shot workloads run a fine-tuned classifier and a zero-shot model on internal H100 and MI300X fleets.

With the same \bound and \ride, the shaper matched demand to capacity in the fixed reservation, the variable on-demand pool, and resized internal fleets. It needed no capacity retuning through nine fleet resizes and an elastic pool's preemptions (Section~\ref{sec:fleet}). It worked alongside engine tuning (Section~\ref{sec:lessons}). Three workloads replaced precursors that queued work durably and, without \ride, dispatched it with stateless jobs at a hand-set global rate. The reserved workload's precursor, the static system, rode the same provider admission as the shaper and differed only in its dispatcher. In a pre-launch test, the shaper took over. Within two hours, refusals fell from about 23\% to 3--5\%, near the shaper's 5\% target. Refusals rose again when the static system resumed. At launch, the reservation's unused capacity more than halved (observational). Since the provider updated its admission, the shaper's utilization has averaged 95.8\% of the reservation and 99.1\% through a backlogged week (Section~\ref{sec:thirdparty}).

For the classifier and zero-shot precursors, the global rate was in effect a \bound blind to engine state. In concurrent A/Bs on production traffic, the shaper, with a \bound at each engine and \ride, served the classifier on 43\% fewer H100s than its precursor. Under the shaper, lower-priority work waited longer at a backfill's peak. On equal zero-shot fleets of about 600 MI300X each, the shaper drained every shared backlog first, by a median 3 hours. When both ran at their dispatch limits, the shaper's engines ran a median 124 of 128 sequences, against the precursor's 41, and served 9\% more per GPU on prefill-heavy work (Section~\ref{sec:fleet}).

In replay without a \bound, the engine stayed busy, but goodput fell to zero. At a stable ceiling, \ride behind per-replica caps came within \rqAbsGap of the goodput of a rate fixed 20\% above measured capacity, with a refusal share \rqRejRatio of the fixed rate's (Sections~\ref{sec:controlled} and~\ref{sec:follow}). A well-tuned fixed rate can match or beat \ride while the ceiling holds, but does not move with the ceiling. If the ceiling doubles, the fixed rate leaves half the capacity idle. If the ceiling halves, half its attempts are refused. The ceiling moves with fleet resizes, elastic-pool preemptions, and token-mix shifts. The precursors' fixed rates were retuned by hand repeatedly.

In production, \ride has worked with four kinds of \bound: model-server caps, provider admission, a per-caller token quota in our gateway, and an engine queue-depth aware router. \system composes established mechanisms above a serving framework, using the framework's per-replica limit as the \bound and adding the durable priority queue and refusal-driven fleet-level rate that serving frameworks lack.

We make four contributions. We present an LLM dispatch design that composes durable priority waiting, fail-fast admission, and feedback-driven dispatch across provider and internal backends, so fleets can be sized for typical load instead of peak. Section~\ref{sec:scaleeng} describes the engineering that makes a simple rule safe on a changing fleet: central placement on a consistent hash ring with hysteresis, one leader that pushes each worker its share of each queue's rate, round-robin service across many queues per worker, drain and freeze through restarts, and runtime knobs for every constant. We report evidence at two levels, controlled replay on real engines (Sections~\ref{sec:controlled} and~\ref{sec:follow}) and production, including same-traffic A/Bs for two workloads (Section~\ref{sec:fleet}). Section~\ref{sec:lessons} draws lessons: cap the engine's queue, use the dispatcher to locate the ceiling, pick the \bound by its job, keep dispatch stateful, make the simple path the only path, and size the fleet for freshness.

\section{Production Setting and Requirements}
\label{sec:setting}

Inference results feed downstream products under freshness goals set by priority, and no user waits on a request. Producers submit work through one enqueue interface. We study its online lane, and Appendix~\apBatch covers the batch lane.

The traffic shaper serves three resource models. Reserved provider capacity is a fixed token allocation, and measured utilization can briefly exceed 100\%. On-demand capacity varies because its quota caps consumption but reserves no hardware, and the provider's other callers share the pool behind that quota. When that pool is congested, reachable capacity falls below quota and the provider sheds the excess with HTTP 429. Internal capacity is vLLM inference engines on H100, MI300X, and MI350X fleets, each wrapped by a custom model server. A replica is one engine (one H100 or two MI300X). The model server's per-replica request cap decides which requests enter that replica. Another model runs on a fourth accelerator type in external data centers. Across the six workloads, fleets held more than 4,000 accelerators at their peaks, plus reserved provider capacity (Appendix~\apScale).

We see a provider only through the 429s it returns for a capacity domain as a whole. Inside a serving pool, \bound refuses at each replica, and the pool admits the sum of its replicas' caps. In production, callers sharing a capacity domain share one online queue and one dispatch rate.

A tenant is a serving deployment on our platform, and an attempt is one dispatch, refused or not. Freshness is enqueue-to-result time, spanning queue wait and inference. The reserved, classifier, and zero-shot workloads had precursors, and we call the reserved workload's precursor the static system. All three precursors ran stateless jobs on a serverless asynchronous job platform built on \foqs. Each job type had one global, hard-coded dispatch rate that engineers set and adjusted by hand. Failed requests retried independently, without feedback.

Adding capacity is slow. A new replica takes a median of 136~s (classifier) to 159~s (zero-shot) to load its model and serve. The shaper's control loop runs every 15 seconds. Even with GPU quota granted, two classifier resizes took 27 and 13 minutes from landing to serving, excluding review. They grew the fleet from 34 to 45 H100s (Appendix~\apAutoscale). \system assumes work that can wait and backends that refuse promptly, with one controller per capacity domain.

\section{Pitfalls Under Shifting Supply and Demand}

Dispatchers that back off on refusals, like \ride, cope with shifting supply and demand (Section~\ref{sec:related}), unlike the designs below.

\paragraph{Static limits.}
A fixed request rate protects a backend only for the workload mix used to choose it. A lighter mix strands capacity, and a heavier one overloads execution. A fixed rate cannot follow moving capacity. Token-aware limits help when token counts are known. But generative output length is unknown at admission. Even fine-tuned predictors estimate that length poorly~\cite{Zhang2026JITServe}.

\paragraph{Queues inside the server.}
Inference engines queue accepted work. By default, vLLM holds requests beyond its running batch in a waiting queue. Orca and vLLM improve how an engine schedules and stores sequences~\cite{Yu2022Orca,Kwon2023PagedAttention}. But neither decides how much asynchronous work waits inside the replica. A model server that limits concurrency by blocking callers adds a second, in-memory queue. Requests in either queue hold memory and connections while their deadlines run, without durable priority or redelivery. Across a fleet, each engine serves and times out its own line, in its own order. A request retried on timeout can land behind a longer line on another replica, so priority becomes best effort. \bound refuses instead of blocking. A refused request returns promptly to the head of its worker's ready queue, outside the engine (Section~\ref{sec:scaleeng}). On lease expiry, the request returns to the shared priority queue, where backlog is visible and recoverable.

\paragraph{Peak provisioning and autoscaling.}
Provisioning for an observed peak cannot cover an arbitrary backfill and holds accelerators for work that could wait. Over three months, serving mean classifier arrivals needed 25 H100s. The busiest hour reached 4.0 times that mean. Serving that hour on arrival would need at least 99 H100s (Appendix~\apAutoscale). Autoscaling alone falls short. At the classifier's sharpest one-hour rise, a fleet that reacts within an hour, without a queue, would need at least 74 more H100s than the preceding hour's arrivals required. Capacity arrives late and leaves abruptly, in whole-replica steps (Section~\ref{sec:ondemand} and Appendix~\apAutoscale). Reserved provider capacity cannot be autoscaled.

\section{Cap-and-Track Design}
\label{sec:design}

\subsection{Traffic shaper overview}

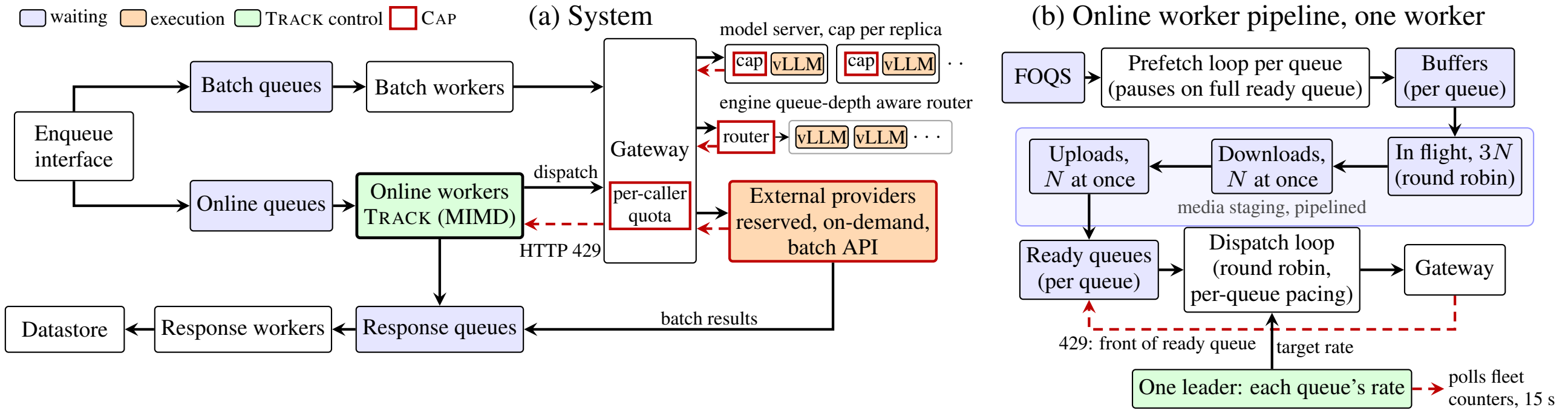
\begin{figure*}[t]
\centering
\begin{tikzpicture}[x=1cm, y=1cm, >=stealth, font=\scriptsize,
  wait/.style={draw, fill=blue!10, rounded corners=1pt, align=center, inner sep=2pt},
  exec/.style={draw, fill=orange!30, rounded corners=1pt, align=center, inner sep=2pt},
  proc/.style={draw, fill=white, rounded corners=1pt, align=center, inner sep=2pt},
  lbl/.style={font=\tiny, inner sep=1pt},
  flow/.style={->, thick},
  fb/.style={->, densely dashed, thick, red!75!black}]
\node[font=\small] at (6.35,3.6) {(a) System};
\node[wait, minimum width=0.28cm, minimum height=0.18cm] (lw) at (0.25,3.6) {};
\node[anchor=west, lbl] at (lw.east) {waiting};
\node[exec, minimum width=0.28cm, minimum height=0.18cm] (le) at (1.35,3.6) {};
\node[anchor=west, lbl] at (le.east) {execution};
\node[proc, fill=green!14, minimum width=0.28cm, minimum height=0.18cm] (lr) at (2.6,3.6) {};
\node[anchor=west, lbl] at (lr.east) {\textsc{Track} control};
\node[draw=red!75!black, line width=0.8pt, fill=white, minimum width=0.28cm, minimum height=0.18cm] (lb) at (4.3,3.6) {};
\node[anchor=west, lbl] at (lb.east) {\textsc{Cap}};
\node[proc, minimum width=1.3cm, minimum height=0.75cm] (call) at (0.7,2.2) {Enqueue\\interface};
\node[wait, minimum width=1.55cm, minimum height=0.55cm] (bq) at (2.75,2.85) {Batch queues};
\node[proc, minimum width=1.6cm, minimum height=0.55cm] (bw) at (4.7,2.85) {Batch workers};
\node[wait, minimum width=1.55cm, minimum height=0.55cm] (oq) at (2.75,1.55) {Online queues};
\node[proc, fill=green!14, minimum width=1.6cm, minimum height=0.7cm, line width=0.9pt] (ow) at (4.7,1.55) {Online workers\\\textsc{Track} (MIMD)};
\node[proc, minimum width=0.85cm, minimum height=2.45cm] (gw) at (7.0,2.15) {Gateway};
\node[draw=red!75!black, line width=0.8pt, fill=white, font=\tiny, align=center, inner sep=1pt] (gq) at (7.0,1.55) {per-caller\\quota};
\node[lbl, anchor=west] at (7.72,3.46) {model server, cap per replica};
\node[font=\tiny] at (10.3,3.1) {$\cdots$};
\foreach \i/\x in {1/8.38,2/9.6} {
  \node[draw, fill=white, rounded corners=1pt, minimum width=1.12cm, minimum height=0.42cm] (e\i) at (\x,3.1) {};
  \node[draw=red!75!black, line width=0.8pt, fill=white, font=\tiny, minimum width=0.3cm, minimum height=0.24cm, inner sep=0.8pt] at (\x-0.29,3.1) {cap};
  \node[exec, minimum width=0.42cm, minimum height=0.24cm, font=\tiny, inner sep=0.3pt] at (\x+0.23,3.1) {vLLM};
}
\node[lbl, anchor=west] at (7.72,2.66) {engine queue-depth aware router};
\node[draw=red!75!black, line width=0.8pt, fill=white, font=\tiny, minimum height=0.34cm, inner sep=1.5pt] (rt) at (8.05,2.3) {router};
\draw[gray!70, rounded corners=1pt] (8.52,2.12) rectangle (10.3,2.48);
\node[font=\tiny] at (10.05,2.3) {$\cdots$};
\foreach \i/\x in {1/8.88,2/9.52} {
  \node[exec, minimum width=0.5cm, minimum height=0.24cm, font=\tiny, inner sep=0.3pt] (r\i) at (\x,2.3) {vLLM};
}
\draw[->, thin] (rt.east) -- (8.52,2.3);
\node[exec, draw=red!75!black, line width=0.8pt, minimum width=2.1cm, minimum height=0.66cm] (ext) at (9.0,1.38) {External providers\\reserved, on-demand,\\batch API};
\node[wait, minimum width=1.6cm, minimum height=0.5cm] (rq) at (4.7,0.2) {Response queues};
\node[proc, minimum width=1.55cm, minimum height=0.5cm] (rw) at (2.55,0.2) {Response workers};
\node[proc, minimum width=1.3cm, minimum height=0.5cm] (ds) at (0.6,0.2) {Datastore};
\draw[flow] (call.north) |- (bq.west);
\draw[flow] (call.south) |- (oq.west);
\draw[flow] (bq) -- (bw);
\draw[flow] (oq) -- (ow);
\draw[flow] (bw.east) -- (gw.west |- bw);
\draw[flow] (ow.east |- 0,1.75) -- (gw.west |- 0,1.75);
\node[lbl] at (6.08,1.9) {dispatch};
\draw[fb] (gw.west |- 0,1.35) -- (ow.east |- 0,1.35);
\node[lbl] at (6.02,1.06) {HTTP 429};
\draw[flow] (gw.east |- 0,3.17) -- (e1.west |- 0,3.17);
\draw[fb] (e1.west |- 0,3.03) -- (gw.east |- 0,3.03);
\draw[flow] (gw.east |- 0,2.4) -- (rt.west |- 0,2.4);
\draw[fb] (rt.west |- 0,2.2) -- (gw.east |- 0,2.2);
\draw[flow] (gw.east |- 0,1.47) -- (ext.west |- 0,1.47);
\draw[fb] (ext.west |- 0,1.3) -- (gw.east |- 0,1.3);
\draw[flow] (ow.south) -- (rq.north);
\draw[flow] (ext.south) |- node[lbl, above, pos=0.7] {batch results} (rq.east);
\draw[flow] (rq) -- (rw);
\draw[flow] (rw) -- (ds);
\begin{scope}[xshift=10.8cm]
\node[font=\small] at (2.85,3.6) {(b) Online worker pipeline, one worker};
\node[wait, minimum width=0.9cm, minimum height=0.55cm] (f) at (0.5,2.95) {FOQS};
\node[proc, minimum width=1.5cm, minimum height=0.55cm] (pf) at (2.6,2.95) {Prefetch loop per queue\\(pauses on full ready queue)};
\node[wait, minimum width=1.3cm, minimum height=0.55cm] (buf) at (5.0,2.95) {Buffers\\(per queue)};
\draw[fill=blue!4, draw=blue!40, rounded corners=2pt] (0.2,1.33) rectangle (5.86,2.4);
\node[lbl, anchor=south, text=black!70] at (3.0,1.38) {media staging, pipelined};
\node[wait, minimum width=1.3cm, minimum height=0.55cm] (slots) at (5.0,1.98) {In flight, $3N$\\(round robin)};
\node[wait, minimum width=1.3cm, minimum height=0.55cm] (dls) at (3.0,1.98) {Downloads,\\$N$ at once};
\node[wait, minimum width=1.3cm, minimum height=0.55cm] (uls) at (1.0,1.98) {Uploads,\\$N$ at once};
\node[wait, minimum width=1.3cm, minimum height=0.55cm] (rdy) at (1.0,0.85) {Ready queues\\(per queue)};
\node[proc, minimum width=1.6cm, minimum height=0.55cm] (dl) at (3.0,0.85) {Dispatch loop\\(round robin,\\per-queue pacing)};
\node[proc, minimum width=1.1cm, minimum height=0.55cm] (g2) at (5.0,0.85) {Gateway};
\node[proc, fill=green!14, minimum width=2.2cm, minimum height=0.42cm] (ld) at (3.0,-0.45) {One leader: each queue's rate};
\draw[flow] (f) -- (pf);
\draw[flow] (pf) -- (buf);
\draw[flow] (buf) -- (slots);
\draw[flow] (slots) -- (dls);
\draw[flow] (dls) -- (uls);
\draw[flow] (uls) -- (rdy);
\draw[flow] (rdy) -- (dl);
\draw[flow] (dl) -- (g2);
\draw[fb] (g2.south) |- (1.0,0.2) -- (rdy.south);
\node[lbl, anchor=north] at (1.75,0.17) {429: front of ready queue};
\draw[flow] (ld.north) -- node[lbl, right, pos=0.35] {target rate} (dl.south);
\draw[fb] (ld.east) -- ++(0.35,0) node[lbl, right, align=left, text=black] {polls fleet\\counters, 15 s};
\end{scope}
\end{tikzpicture}
\caption{The traffic shaper implements \system. (a) One enqueue interface feeds online and batch queues. \ride runs in the online workers and dispatches each queue through the gateway at its learned rate. A refusal from any of the four kinds of \bound (model-server cap, provider admission, per-caller quota, or engine queue-depth aware router) returns as HTTP 429. Other failures stay outside the feedback loop.  (b) Each worker paces each queue at the rate the leader pushes. A 429 returns the request to the front of its ready queue with media staged. Up to $3N$ media items are in flight, but only $N$ download and $N$ upload at once. The rest wait between the two stages to keep each worker container's network bandwidth saturated.}
\label{fig:architecture}
\end{figure*}

Producers enqueue work into \foqs (Figure~\ref{fig:architecture}a). \ride workers lease requests from \foqs, stage any input media, and dispatch through a gateway to an external provider or an internal model server.

\subsection{Cap: fail fast near execution}\label{sec:bound}

A labeled refusal at the constrained resource exposes overload before a wasteful wait, so the dispatcher needs no model of inference-engine internals or provider accounting.

Providers already return such refusals as HTTP 429. The traffic shaper records each 429, slows dispatch, and keeps the work. Refusals use no reserved tokens.

Internal serving produces the same signal. Before a request reaches vLLM, the model server checks the replica's in-flight count and, at the cap, returns an overload exception that the RPC layer marks as an overload. The gateway maps only overload statuses to HTTP 429 (Appendix~\apRefusal). Without a refusal path, an engine queues work until timeout (Section~\ref{sec:nocap}), so each replica needed its own refusal. We enabled the model server's existing per-replica cap and tied it to engine tuning. Our classifier and zero-shot deployments set vLLM's \texttt{max\_num\_seqs} to 128 and the server's request and processing caps to 140 per replica (together, ``the cap''). This slight oversubscription keeps a few requests waiting in the model server or behind a nearly full batch, so a freed slot refills without a dispatcher round trip (Appendix~\apPercap). Profiling sweeps engine settings to find a workload's highest-throughput configuration, including the concurrency where throughput plateaus. Set the cap with the engine settings, from profiling or a standard deployment template for a token mix in a validated range. The cap needs no precise tuning. In replay, caps at two and four times the plateau kept goodput within 2.4\% of plateau goodput. There, requests above the plateau still queue in the engine, and latency rises two- to fourfold. In replay, a per-replica token budget cut the classifier's tail latency but gave no clear gain on a generative workload, so we leave that token budget to future work (Appendix~\apBudget).

\ride needs only a timely, labeled refusal and no engine metrics. In production, \ride worked behind four kinds of \bound without knowing how they decide (Appendix~\apBounds). Model-server caps refuse on per-engine load and need no size prediction. Provider admission refuses up front. Its ceiling moved with our token mix, as if the provider predicted each request's size (Section~\ref{sec:thirdparty}). Our gateway's per-caller quota meters tokens at response time, so it admits past its token budget before refusals spike (Section~\ref{sec:scale}). Response-time metering accounts exactly for requests whose output length is unknown up front, but the quota's cyclic signal makes \ride's rate swing. Like the caps, the engine queue-depth aware router (below) refuses on per-engine load. The precursors' hand-set global limit, the one production \bound that ran without \ride, saw no engine state.

Each model-server cap covers one replica. Fleet-level admission is the sum over live replicas, so it rises and falls as the fleet resizes, restarts, or loses hosts, with no reconfiguration (Section~\ref{sec:fleet}). On internal fleets, the RPC layer's router sends each request to the less loaded of two sampled replicas. Skewed load can still make a replica refuse while others have headroom. \ride reads that refusal as a lower ceiling. In the zero-shot A/B, skew left little capacity idle. In hours when dispatch was capped, every region with at least nine engines had 89--100\% of its engines running at least 120 of 128 sequences (Appendix~\apPercap). An engine queue-depth aware router reduces that skew and moves refusals from the engines to itself. This router sees each engine's running and waiting work and routes to engines with room. It refuses once an engine's queue passes its limit (Appendix~\apBounds). The router adds a central request-path component with a lagging per-engine view. It could also target engines tuned for characteristics like sequence length, which we have not tested.

\subsection{Track: follow supply}\label{sec:ride}

Every 15 seconds, a leader worker (Section~\ref{sec:scaleeng}) sets each queue's dispatch rate from that queue's refusal share, the fraction of fleet-wide successes and 429s that are 429s. At or below our target refusal share (2\% in production), the leader raises the rate 3\%. Above the target, the leader cuts the rate in three tiers (Appendix Figure~\apFigAimd). For shares up to 50\%, the multiplier is 0.99. Above 50\% and up to 65\%, it is 0.85. Above 65\%, it is 0.5. Every queue uses these constants and learns its rate from its own refusal share. The controller is multiplicative in both directions (MIMD), and the target only switches the rate from rising to falling.

We began with additive-increase, multiplicative-decrease (AIMD) control, but its fixed step overshot low-rate queues and was too slow for high-rate ones (Appendix~\apCtrl). With one controller per capacity domain, \ride needs none of AIMD's fairness between senders. One 3\% increase takes about three 0.99 cuts to undo, so the rate traces a small sawtooth around the ceiling. The refusal share hovers above the target, near 4--5\% behind provider admission and, in the classifier backfill, near 6--8\% behind per-replica caps (Appendix~\apScale). A higher target keeps throughput closer to the ceiling but draws more refusals. The mild tier keeps swings small, and harder tiers act only when more than half of attempts are refused. \ride responds slowest to moderate drops. Losing 40\% of capacity draws about 40\% refusals, still the mild tier. Reaching the post-loss capacity takes about 51 decisions (13 minutes). This slow response adds refusals, but each returns fast and the work stays queued (Section~\ref{sec:follow}). Section~\ref{sec:ondemand} shows the climb back.

In production, the shaper has run \ride at up to about 16,000 requests per second through nine fleet resizes. In one week, the rate fell 30\% or more in only 14 episodes. Even behind the per-caller quota, whose metering lags by an inference time, \ride's refusals tracked the quota's blocks minute by minute, with timeouts in one minute of the week (Appendices~\apCtrl, \apAutoscale, \apBounds, and \apSafe).

\subsection{Priority and waiting}

 An integer priority score orders the queue and selects the lane. Higher-priority work consumes capacity first, and lower-priority work waits. Ties dequeue roughly FIFO. By product choice, waiting items gain no priority with age. A caller can give each item a time to live, checked before dispatch, bounding the item's wait under strict priority. Over one month, the classifier's two lowest-priority bands met their freshness goals for 69--83\% of items (Appendix~\apFresh).

\section{Running a Simple Rule at Scale}
\label{sec:scaleeng}
\label{sec:lifecycle}

Running \ride's simple rule on a changing fleet took most of the traffic shaper's engineering. Most mechanisms below fixed problems we hit (Figure~\ref{fig:architecture}b and Appendix~\apCoord). None is LLM-specific, but LLM serving's shifting supply and demand made them necessary.

\paragraph{One controller per capacity domain.} Our first prototype ran a controller in each of $N$ workers, making $N$ independent sawtooths on one quota. The worker with the lexicographically smallest hostname in fleet discovery leads, without a consensus service. Every 15 seconds the leader pools every worker's counts, decides each queue's fleet-wide rate, and pushes each worker its share. A brief two-leader overlap only mis-sets a rate until the overlap ends, while caps or provider 429s refuse excess. A new leader seeds each rate from workers' highest reported targets.

\paragraph{Placement sized for the signal.} Spreading a low-rate queue over all workers gave each worker too few dispatches per poll to estimate the queue's refusal share reliably. The leader places each queue on just enough workers for its rate. A consistent hash ring picks those workers, so a worker joining or leaving moves about $1/N$ of placements. A queue's worker count changes only when the new count differs by two or more. This hysteresis keeps the sawtooth from churning placement.

\paragraph{Only capacity refusals count.} A timeout can mean a request waited or ran past its deadline on a busy inference engine, where backing off helps. It can also mean network delays or another infrastructure fault, where backing off only idles capacity. A dispatcher cannot tell timeouts apart, so only the \bound's labeled 429s feed the rate. An early \ride controller counted timeouts. A burst of them from network delays cut dispatch and utilization while work stayed queued. We accept extra network load to prioritize keeping accelerators busy.

\paragraph{No climbing in quiet hours.} Overnight demand troughs brought no refusals, so early versions let targets climb toward their maximum for days. Once work returned, the inflated targets overran the backend. A utilization gate now blocks increases when dispatch is below 75\% of the target rate (Appendix Figure~\apFigAimd).

\paragraph{Restarts are not capacity changes.} During a rolling restart, a draining worker and its replacement can dispatch at once. Their burst of 429s can push the refusal share into the controller's harder cut tiers. A restart proceeds in steps over hours, so these cuts would repeat at every step. The leader freezes rates until 10 polls after the last worker-count change. A brief refusal spike does little harm, because refused work stays queued. Each needless cut would idle capacity for minutes as \ride climbs back. Until a freeze lifts between rollout steps, rates ignore capacity changes and the \bound refuses excess. Across 31 rolling deploys, the reserved workload's throughput was a median 1.001 times its pre-deploy level. But a frozen controller cannot back off. One worker missing a poll during, say, a garbage-collection pause or network failure could freeze every queue's rate. Scaling the fleet for more load made such misses more likely, so we added hysteresis. A freeze now needs a drop of at least 10\% of workers. \foqs redelivers a dead worker's unacknowledged items once leases expire, so delivery is at least once unless storing a response fails.

\paragraph{Retries stay local.} Returning refused requests to the shared queue would bounce them across workers. Each bounce would restage media until retry limits sent the requests to a dead-letter queue. A refused request now returns to the head of its worker's ready queue, with media staged. Only lease expiry returns the request to \foqs. A worker dispatches only when the remaining lease exceeds the request timeout, so no request is still running at lease expiry when \foqs redelivers it.

\paragraph{Many queues, any backend.} Each worker serves its queues round robin. Each queue has a fetch loop, rate-scaled prefetch depth, and ready queue, so one queue's backlog cannot stall another (Section~\ref{sec:ondemand}). Media streams to temporary files on local disk. Request attributes name the queue, created on its first write. A new caller needs only configuration and a cap, from profiling or a standard deployment template. A new backend needs only an adapter returning capacity refusals as HTTP 429.

\paragraph{Tuning live.} Every controller constant is a runtime knob, so we tuned the cut tiers on production traffic without a deploy. Changing the step and target refusal share live tests whether the backend would admit more, because they shape how far the sawtooth overshoots the ceiling before each cut. If neither more refusals nor a different rate shape moves utilization, the ceiling sits in the backend's admission (Section~\ref{sec:thirdparty}). A \bound and \ride can be tuned together, whoever owns the \bound.

\section{Evaluation}
\label{sec:evaluation}

 Section~\ref{sec:collapse} tests whether the \bound prevents collapse. Section~\ref{sec:follow} tests whether \ride matches a tuned fixed rate at a stable ceiling and follows a moving ceiling. Section~\ref{sec:fleet} measures fleet size and freshness in production. Section~\ref{sec:scale} tests \ride across backends at scale. Replay on real inference engines isolates the \bound and compares \ride with fixed rates. The two same-traffic A/Bs are the controlled production comparisons. Other windows, where post volume, media size, token mix, and routing vary daily, are observational consistency checks.

\begin{table*}[t]
\centering
\small
\setlength{\aboverulesep}{0.15ex}\setlength{\belowrulesep}{0.25ex}
\caption{Main production results and their caveats. In reserved rows, utilization is the share of the reservation used, including overnight troughs. Refused is the share of attempts drawing 429s. The busiest quarter is the quarter of hours with the most attempts.}
\label{tab:results}
\begin{tabular}{>{\raggedright\arraybackslash}p{0.12\textwidth}>{\raggedright\arraybackslash}p{0.26\textwidth}>{\raggedright\arraybackslash}p{0.34\textwidth}>{\raggedright\arraybackslash}p{0.16\textwidth}}
\toprule
\textbf{Setting} & \textbf{Compared (window)} & \textbf{Result} & \textbf{Caveat} \\
\midrule
On-demand capacity loss & 26 h of loss (three mornings) against median rates & Other three workloads at 98.0--99.2\%. On-demand queue emptied 2--3 h after recovery & Hourly, observational \\
\midrule
Reserved, launch & Shaper's first 20 days, target refusal share raised 5\% to 30\% in 2 days, against static system's prior 23 days & Utilization 94.8\% (first 4 full days) against 88.2\%. Refused 28.3\% (first 4 days) and 27.6\% (20 days), against 16\% (24.4\% in the busiest quarter of hours) & Observational. Ceiling near 95--97\% \\
\midrule
Reserved, steady state & Shaper only, 15 weeks after the provider updated its admission & Utilization 95.8\% (six-hour periods: median 98.6\%, top-quarter mean 99.6\%). Refused 4.2\% in periods with 429s, 4.4\% in the busiest quarter of hours & No baseline after the update \\
\midrule
Classifier A/B, fleet sizing & Shaper against precursor on full traffic. Arrivals over 3 months & Precursor 68 H100s, erroring at about 2/3 of its profiled per-card rate. Shaper 39 at equal coverage and scores, 45 in production & Serve-on-arrival fleets (Appendix~\apAutoscale) are lower bounds \\
\midrule
Zero-shot, equal fleets & Shaper against precursor, 23 days, about 600 MI300X each & Shaper drained all shared backlogs first, a median 3 h sooner. At both paths' dispatch limits, its engines ran a median 124 of 128 sequences against the precursor's 41, and served 9\% more per GPU & Prefill-heavy model. Fewer-GPU estimate untested \\
\bottomrule
\end{tabular}
\end{table*}

\subsection{The \bound prevents collapse}
\label{sec:collapse}
\label{sec:controlled}

A per-replica cap kept excess work waiting in the durable queue instead of timing out inside execution. Without a \bound, goodput collapsed in replay, and timeouts piled up in production. We replayed classifier and generative request shapes on one A100 behind the production model server and gateway. Each ran at about twice measured capacity, with and without a cap where replica throughput plateaued (Appendix~\apReplay).

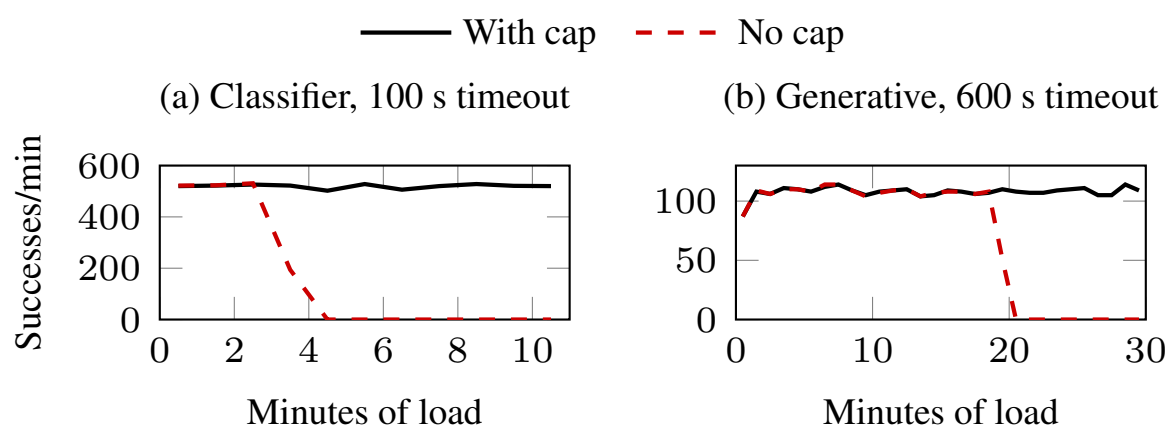
\begin{figure}[t]
\centering
\begin{tikzpicture}
\begin{axis}[name=cl, width=0.52\columnwidth, height=2.6cm, xlabel={Minutes of load}, ylabel={Successes/min},
  ymin=0, ymax=600, xmin=0, xmax=11, xtick={0,2,4,6,8,10}, tick label style={font=\scriptsize}, label style={font=\scriptsize},
  title style={font=\scriptsize}, title={(a) Classifier, 100 s timeout}]
\addplot[thick, black, mark=none] coordinates {(0.5,520) (1.5,522) (2.5,526) (3.5,522) (4.5,502) (5.5,528) (6.5,506) (7.5,520) (8.5,528) (9.5,521) (10.5,520)};\label{nocap:with}
\addplot[thick, red!80!black, dashed, mark=none] coordinates {(0.5,522) (1.5,523) (2.5,531) (3.5,194) (4.5,0) (5.5,0) (6.5,0) (7.5,0) (8.5,0) (9.5,0) (10.5,0)};\label{nocap:no}
\end{axis}
\begin{axis}[name=gen, at={(cl.east)}, anchor=west, xshift=1.1cm, width=0.52\columnwidth, height=2.6cm, xlabel={Minutes of load},
  ymin=0, ymax=130, xmin=0, xmax=30, xtick={0,10,20,30}, tick label style={font=\scriptsize}, label style={font=\scriptsize},
  title style={font=\scriptsize}, title={(b) Generative, 600 s timeout}]
\addplot[thick, black, mark=none] coordinates {(0.5,87) (1.5,108) (2.5,106) (3.5,111) (4.5,110) (5.5,108) (6.5,112) (7.5,114) (8.5,109) (9.5,105) (10.5,108) (11.5,109) (12.5,110) (13.5,104) (14.5,105) (15.5,109) (16.5,108) (17.5,106) (18.5,107) (19.5,110) (20.5,108) (21.5,107) (22.5,107) (23.5,109) (24.5,110) (25.5,111) (26.5,105) (27.5,105) (28.5,114) (29.5,109)};
\addplot[thick, red!80!black, dashed, mark=none] coordinates {(0.5,87) (1.5,109) (2.5,106) (3.5,110) (4.5,110) (5.5,108) (6.5,114) (7.5,114) (8.5,109) (9.5,104) (10.5,107) (11.5,109) (12.5,110) (13.5,104) (14.5,107) (15.5,108) (16.5,108) (17.5,106) (18.5,108) (19.5,51) (20.5,0) (21.5,0) (22.5,0) (23.5,0) (24.5,0) (25.5,0) (26.5,0) (27.5,0) (28.5,0) (29.5,0)};
\end{axis}
\path (cl.outer north) -- (gen.outer north) node[midway, above=-1pt, font=\scriptsize] {\ref{nocap:with}~With cap\quad \ref{nocap:no}~No cap};
\end{tikzpicture}
\caption{Controlled replay at about twice a replica's capacity, median of three runs per arm. Without the cap, goodput (successes per minute) falls to zero and stays there once requests queued inside the model server outlive the timeout. With the cap, goodput holds steady.}
\label{fig:nocapgen}
\end{figure}

Without the cap, the engine stayed busy at the capped arm's token rate, but goodput fell to zero in both shapes and never recovered (Figure~\ref{fig:nocapgen} and Appendix~\apCodel). Collapse began once requests queued in the model server outlived the timeout. An engine that cancellation does not reach keeps serving requests whose callers have left~\cite{Bronson2021Metastable,Huang2022Metastable}. With the cap, goodput held steady. Excess requests waited in the dispatcher's queue, outside the timeout clock.

Of the replayed fixes, only refusing restored goodput. Dropping or canceling before vLLM did not. A CoDel-style rule, itself a \bound that refuses on queue delay, matched the cap's goodput (Appendix~\apCodel). \ride does not replace the \bound. We ramped a \ride variant that counts timeouts as refusals into an uncapped engine. Under every increase rule, the variant reached about a third less goodput than behind the \bound (Appendix~\apRamp).

\label{sec:nocap}
The cap ended the classifier's production timeouts. Uncapped replicas held excess work. Per-minute p99 latency, averaged hourly, climbed from 5~s to 511~s (Figure~A1 and Appendix~\apFleetsrc). After timeouts recurred twice, caps were set to 140 per replica. Once refusals began, the next six days had no timeouts. The zero-shot workload followed the same sequence (Appendix~\apEpisodes).

\label{sec:capped}
During the four-day classifier backfill, priority kept most high-priority work fresh. Excess work waited in \foqs instead of engines (Appendix~\apEpisodes, Figure~A1). Over the backfill's last three days, the top band met its 30-minute goal for 99\% of items. The second-highest band met its 3-hour goal for 89\%. Lower-priority work waited longer at peaks on the shaper's classifier fleet, which was smaller than the precursor's. The two lowest bands met their goals for 44--55\% of items (Section~\ref{sec:fleet}, Appendix~\apFresh).

\subsection{\ride follows a moving ceiling}
\label{sec:follow}

In replay on real inference engines, through a production capacity loss, and on a provider reservation, \ride served about as much as a tuned fixed rate at a stable ceiling and followed the ceiling when it moved.

A second replay held caps and queue constant and varied only dispatch (Figure~\ref{fig:controlled}, Appendix~\apRevfixed). On each of three hosts, one of two replicas was removed for 10 minutes and restored. At a per-replica cap of 40, a small queue forms in vLLM behind a full batch. About two batches wait at 64. Fixed-rate arms used the same caps without \ride, at 50, 100, or 120\% of capacity measured beforehand. \ride started at half capacity and was told nothing else.

At a stable ceiling, \ride served \rqRideFullG requests per second per host, slightly more than the 100\% rate (\rqFixFullG\rqFixSrc) and about as much as the 120\% rate (\rqHighFullG). Because \ride probes for the ceiling, it drew refusals on \rqRideFullShare of attempts. The 100\% rate drew \rqFixFullShare and the 120\% rate \rqHighFullShare.

With one replica out, the 100 and 120\% rates drew refusals on \rqFixCutShare and \rqHighCutShare of attempts, but \ride's refusals fell as it descended toward the new ceiling. After the restore, the 100 and 120\% rates returned at once, and \ride regained full capacity within about 5 minutes. The lowest fixed rate stayed at cut capacity, half of full capacity. Refusal share matters because refused attempts still load the gateway, network, and provider or internal-tier admission, and media stays staged until a retry succeeds.

Production is closest to the small-queue replay setting. Request caps sit just above the batch limit (140 against 128 sequences), and batches run nearly full (median 124 running, Section~\ref{sec:fleet}). With so short a queue, a less precise dispatcher can loosen the \bound to build a longer queue and keep batches full. It can also send above capacity like the 120\% fixed rate, accepting more refusals and latency (Appendix~\apRevfixed).

In emulation, before a capacity drop, classic AIMD served about 66\% of capacity against about 96\% for \ride. A simulation ran with capacity scaled 25-fold. At a stable ceiling, the 10\% step's target rate swung 0.222 of capacity, against 0.065 for \ride (Appendix~\apCtrl). A larger step recovers faster after deep drops but widens the sawtooth at the ceiling, adding refusals. One 3\% step served every backend (Appendix~\apSafe).

\input{figures/fig_controlled}

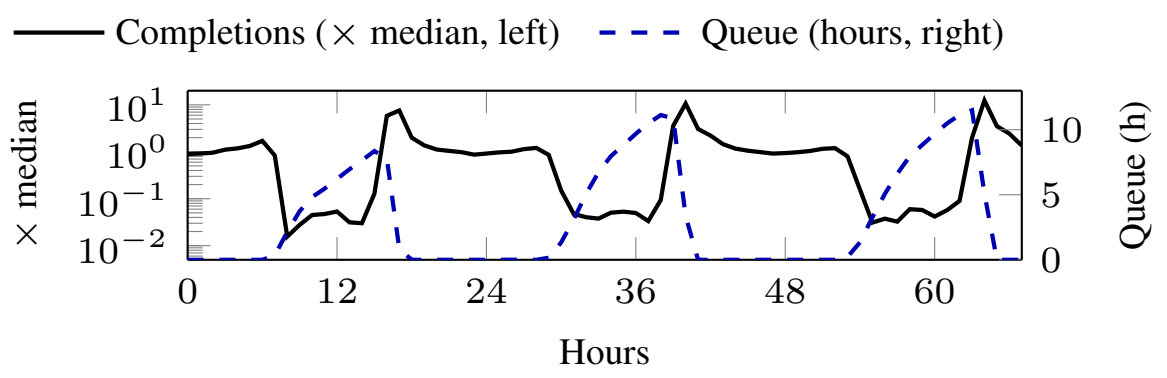
\begin{figure}[t]
\centering
\begin{tikzpicture}
\begin{axis}[name=cl4, width=0.86\columnwidth, height=2.7cm, xlabel={Hours}, ylabel={$\times$ median},
  ymode=log, ymin=0.005, ymax=20, xmin=0, xmax=67, xtick={0,12,24,36,48,60},
  axis y line*=left, tick label style={font=\scriptsize}, label style={font=\scriptsize}]
\addplot[thick, black] coordinates {(0,0.8941) (1,0.9167) (2,0.9541) (3,1.1079) (4,1.1822) (5,1.3305) (6,1.7112) (7,0.8332) (8,0.0156) (9,0.0278) (10,0.0450) (11,0.0472) (12,0.0534) (13,0.0314) (14,0.0301) (15,0.1278) (16,5.8374) (17,7.5738) (18,1.9982) (19,1.3700) (20,1.1186) (21,1.0446) (22,0.9810) (23,0.8692) (24,0.9165) (25,0.9703) (26,1.0029) (27,1.1380) (28,1.2092) (29,0.8616) (30,0.1502) (31,0.0476) (32,0.0404) (33,0.0377) (34,0.0504) (35,0.0530) (36,0.0496) (37,0.0334) (38,0.0939) (39,3.5664) (40,10.6418) (41,3.0566) (42,2.2121) (43,1.4511) (44,1.1688) (45,1.0531) (46,0.9826) (47,0.9155) (48,0.9388) (49,0.9782) (50,1.0402) (51,1.1579) (52,1.1941) (53,0.8010) (54,0.1571) (55,0.0310) (56,0.0376) (57,0.0325) (58,0.0598) (59,0.0576) (60,0.0417) (61,0.0571) (62,0.0897) (63,1.9822) (64,12.3945) (65,3.5119) (66,2.4374) (67,1.3554)};\label{pl:comp}\label{cl:comp}
\end{axis}
\begin{axis}[width=0.86\columnwidth, height=2.7cm, ylabel={Queue (h)}, ymin=0, ymax=13, xmin=0, xmax=67,
  axis y line*=right, axis x line=none, tick label style={font=\scriptsize}, label style={font=\scriptsize}]
\addplot[dashed, blue!70!black, thick] coordinates {(0,0.00) (1,0.00) (2,0.00) (3,0.00) (4,0.00) (5,0.00) (6,0.00) (7,0.34) (8,2.14) (9,3.74) (10,4.82) (11,5.47) (12,6.22) (13,6.96) (14,7.64) (15,8.36) (16,7.79) (17,0.84) (18,0.00) (19,0.00) (20,0.00) (21,0.00) (22,0.00) (23,0.00) (24,0.00) (25,0.00) (26,0.01) (27,0.00) (28,0.00) (29,0.16) (30,1.33) (31,3.09) (32,5.06) (33,6.73) (34,7.97) (35,8.82) (36,9.69) (37,10.49) (38,11.13) (39,10.87) (40,3.36) (41,0.00) (42,0.00) (43,0.00) (44,0.00) (45,0.00) (46,0.00) (47,0.00) (48,0.00) (49,0.00) (50,0.00) (51,0.00) (52,0.00) (53,0.22) (54,1.34) (55,3.26) (56,5.07) (57,6.61) (58,7.96) (59,8.83) (60,9.68) (61,10.49) (62,11.14) (63,11.50) (64,5.09) (65,0.00) (66,0.00) (67,0.00)};\label{cl:queue}
\end{axis}
\node[anchor=south, font=\scriptsize] at (cl4.outer north) {\ref{cl:comp}~Completions ($\times$ median, left)\quad \ref{cl:queue}~Queue (hours, right)};
\end{tikzpicture}
\caption{On-demand workload over the three loss days, hourly from the start of the first. Completions are multiples of the median on the day before the first loss (log scale). The queue is in hours of median completions. Each morning the pool lost capacity and the queue held the excess. Each afternoon \ride followed the pool's 429s back up and drained the queue.}
\label{fig:capacityloss}
\end{figure}

\label{sec:ondemand}
In production, \ride carried the on-demand workload through the loss of most of its capacity and back with no capacity retuning. The other three workloads on the same shaper workers completed 98.0--99.2\% of their median rates (Table~\ref{tab:results}). On three consecutive mornings, on-demand completions fell to a few percent of their pre-loss median. \ride saw this involuntary scale-down only through completions and 429s, so it cut its rate to its floor while work queued. Each afternoon \ride raised its rate on its own, and the queue emptied within two to three hours (Figure~\ref{fig:capacityloss}, Appendix~\apAutoscale).

\label{sec:thirdparty}
The provider decides how much of the reservation it admits. The provider's admission acts as the \bound. The ceiling moves with token mix, the input tokens per output token (Appendix~\apBounds). That ratio rose from about 2.9 to 3.5 over four months as teams changed prompts. The reservation stays committed even when unused. Utilization here is the share of the token allocation used and does not measure batch fullness. Internal classifiers are not metered this way.

In the observational pre-launch test, the shaper drew fewer refusals than the static system. The two shared provider admission and differed only in dispatch. The static system's hard-coded rate was set high enough to fill the reservation under any token mix, so the system stayed near the ceiling only by overshooting it. Within two hours of the shaper's takeover, refusals fell from about 23\% to 3--5\%. While the shaper ran alone, an unintended upstream change made the token mix more output-heavy, and the shaper held more of the reservation. Refusals rose again when the static system resumed (Appendix~\apDialup).

Less of the reservation went unused at launch. Over its first four full metered days, the shaper averaged 94.8\% utilization. The static system averaged 88.2\% over the preceding 23 days, under the same provider admission and token mix (Appendix~\apReserved). This observational comparison has no control for backlog. Utilization tracks backlog, and short-queue hours fell in the evening and at night (Appendix~\apBacklog).

More probing did not lift utilization, so the provider's admission set the ceiling. Within two days of launch, we raised \ride's global target refusal share from 5\% to 30\% (Sections~\ref{sec:ride} and~\ref{sec:scaleeng}). The reserved workload was then the shaper's only workload. Refusals rose and the dispatch rate oscillated, but utilization averaged 93--95\% under the usual token mix. These refusals were intended and used no reserved tokens. The output-heavy pre-launch token mix had allowed a higher ceiling (Appendices~\apDialup and~\apReserved).

After the provider's admission update, the shaper followed the higher ceiling with no capacity setting changed. It averaged 95.8\% of the reservation over the next 15 weeks, against 99.1\% over a backlogged week. Overnight hours with an empty online queue account for the gap. The token mix was more input-heavy than at launch. Utilization rose even with a lower target refusal share. The queue absorbed the daily swing in arrivals and kept completions steadier (Appendix~\apBacklog). About 4--5\% of the shaper's attempts were refused, against 16\% (24.4\% in the busiest quarter of hours) under the static system before launch (Table~\ref{tab:results}). There is no baseline after the update (Appendix~\apBacklog).
\subsection{Fleet size and freshness in production}
\label{sec:fleet}

In the same-traffic A/Bs, the shaper served the classifier on 43\% fewer H100s than its precursor and drained zero-shot backlogs first on equal fleets. Because every arm drained a durable queue, each fleet was sized for average load plus spare capacity to drain backlogs in time. None was sized for the peak of every backfill and surge. With per-replica caps, admission followed the growing zero-shot fleet without reconfiguration. During the period of the elastic pool's preemptions, 0.0004\% of requests timed out (Appendix~\apAutoscale).

\paragraph{Classifier.} The shaper served the classifier on 39 cards against the precursor's 68, with coverage and scores at parity. The arms ran concurrently on the same traffic, model, prompts, and vLLM settings. The shaper's modest backlog drained after peaks (Appendix~\apFleetsrc). Profiling sized the precursor fleet. But without a \bound at its inference engines, the precursor began to error just above its usual online load, at about two-thirds of the profiled per-card rate. Of the 29 fewer cards, 5 are the precursor's margin. The rest come from engines at their throughput plateau behind the caps and \ride (Appendix~\apFleetsrc).

Classifier latency matched the precursor's except on the four-day backfill's two peak days. On every other compared day, the shaper's end-to-end latency was no more than 4\% higher than the precursor's in any band at p50, p95, or p99. The median day was at parity (Appendix~\apFleetsrc). On the peak days, the shaper traded freshness for its smaller fleet. Priority kept the shaper's top band fastest, while its lowest band waited a median 6.6 hours in the classifier stage against the precursor's half hour. Each shaper card served about 1.6 times the rate where a precursor card began to error. An equal-sized shaper fleet would drain a backlog about 1.6 times as fast as the precursor, an untested estimate.

\paragraph{Zero-shot.} On equal fleets, the shaper drained every shared backlog first, by a median 3 hours. In the A/B, both paths served every eligible post from separate durable queues on about 600 MI300X each. We shrank neither fleet (Appendix~\apFleetsrc). In peak hours, when either queue's oldest item was over an hour old, the shaper was faster in every band, queue wait included. In the top three bands, its median was about 8~s against the precursor's 2 minutes. Off peak both took seconds, and the shaper was slower, mostly from platform steps before its dispatcher (Appendix~\apFleetsrc).

The shaper's fuller engines served more per GPU. In hours when both paths hit their dispatch limits, its engines ran nearly full batches and served about 9\% more model requests per GPU. Their median was 124 of 128 sequences against the precursor's 41, behind a cap only 12 requests above the batch limit. By an untested estimate, the precursor's daily load would fit on about 48 fewer GPUs, with lower-priority work waiting longer at peaks (Appendix~\apPercap). The gain is modest because the workload is prefill-heavy (about 17 input tokens per output token). We expect decode-heavy work to gain more. The precursor's fixed rate could not follow the post mix. When binding, the hand-retuned rate ran at a median 86\% of the peak request rate both fleets reached, because the work behind each dispatch shifted with that mix through the day (Appendix~\apPercap).

\subsection{Scale across backends}
\label{sec:scale}

The shaper's \ride served several backends behind different kinds of \bound at once. Two workloads joined during a one-day production load test, one on MI350X GPUs, another model on a fourth accelerator type. Four days earlier, one workload's owner cut per-request output and reasoning tokens nearly five-fold. No controller setting changed that week. Test-day permanent errors stayed at or below 0.06\% of completions on every workload (Appendix~\apScale).

The gateway's per-caller quota set the fourth-accelerator model's ceiling. Charging tokens only as responses return, the quota lags usage by an inference time and admits work past its compute budget before refusing. Near the quota, a median 14\% of attempts were refused, against 4--5\% at the reserved workload's ceiling. The gateway attributed these refusals to the quota. Almost no requests waited at the inference engines, and the chips had idle time (Appendix~\apBounds).

During the classifier backfill, three backlogged workloads behind two kinds of \bound refused at once for most of two days (Appendix~\apScale). Online load has since grown to more than 300 million daily requests. Its busiest 24 hours carried 1.1 billion during a backfill (Appendix~\apScale).

\section{Lessons}
\label{sec:lessons}

\ride's rule is simple, but running \ride on a changing fleet took most of the shaper's engineering (Section~\ref{sec:scaleeng}).

\smallskip\noindent\textbf{Cap the engine's queue.}
Turn on the inference engine's refusal path. The vLLM version we deployed limited running sequences but not its waiting queue (Appendix~\apBudget). The model server's request cap was off by default. A cap just above \texttt{max\_num\_seqs} turns waiting inside the engine into 429s that \ride can act on.

Keep little work waiting at the engine. A request held there keeps a connection open through every upstream tier. A dropped link or fleet-update restart can discard the request's engine work and force a retry. Take 5,000 requests per second, about our busiest hourly rate on a typical day. By Little's law, if requests may wait out a 5-minute timeout, up to about 1.5 million are in flight. A 13-second inference keeps about 65,000. Graceful drains of in-flight requests take longer the more work waits.

Set batch size, the \bound's waiting limit, and the timeout together. In staging, raising the batch limit from 256 to 320 sequences added 10--15\% throughput. But with up to 300 requests waiting behind that batch, too many timed out. A longer timeout could have covered that wait. We instead lowered the waiting limit from 300 to 100. The same requests finished sooner, and timed-out attempts fell from 15.4\% to 1.6\% (Appendix~\apBounds).

\smallskip\noindent\textbf{Use the dispatcher to locate the ceiling.}
If utilization stalls while refusals rise, suspect the backend's admission. On the reservation, raising \ride's target refusal share without a deploy showed the provider's admission set the ceiling (Section~\ref{sec:thirdparty}).

\smallskip\noindent\textbf{Pick the \bound by its job.}
Global token budgets, such as our gateway's per-caller quota, serve provisioning, accounting, and fairness, but see no single engine. Per-engine \bound{}s run hardware at its plateau (Section~\ref{sec:bound}). \ride follows any \bound that refuses.

\smallskip\noindent\textbf{Keep dispatch stateful.}
The precursors' stateless jobs (Section~\ref{sec:setting}) were mostly configuration. No invocation remembered the rate or recent refusals. Following the ceiling through refusals requires recent outcomes and the current rate. Stateless jobs need a component that holds this state and adjusts dispatch. In the shaper's long-lived fleet, one leader holds a rate per queue, the last poll's counters, and a freeze counter. A new leader seeds rates from worker reports and starts frozen.

\smallskip\noindent\textbf{Make the simple path the only path.}
Online load roughly doubled in about a month, to more than 300 million requests a day (Appendix~\apScale). \ride has no per-workload capacity setting. Limits come from the \bound's refusals or scale with the fleet through per-replica caps. No caller creates or pre-provisions a queue. The shaper derives one per backend destination from request attributes. Callers that share a model pool share one queue, so no ceiling is split across competing controllers (Section~\ref{sec:scaleeng}). Workers are interchangeable and placed by consistent hashing, so the fleet grew with load without coordination work. Each controller constant (target refusal share, step, freeze) is a runtime knob changed without a deploy. These knobs are shared \ride defaults on every backend and never tuned per workload. The shaper needed no capacity retuning through nine fleet resizes and an elastic pool's preemptions. Its workers served different backends and kinds of \bound at once (Sections~\ref{sec:fleet} and~\ref{sec:scale}).

\smallskip\noindent\textbf{Size the fleet for freshness.}
Under \ride, the fleet runs at its ceiling while work waits in durable queues. The backlog shows how far demand has outrun supply. Fleet size no longer protects against overload. Chosen from freshness goals, work priority, and GPU availability, it sets how fresh lower-priority work stays. In the A/B, the shaper served the classifier's traffic on 39 H100s. The two lowest-priority bands met their freshness goals for 44--55\% of items over a backfill's last three days (Section~\ref{sec:capped}). The classifier now runs on 45 H100s to drain peaks faster.

\section{Related Work}
\label{sec:related}

\smallskip\noindent\textbf{Overload and admission control.}
Overload control drops or delays excess work. SEDA acts at stage boundaries~\cite{Welsh2001SEDA}, DAGOR by business priority~\cite{Zhou2018DAGOR}. Breakwater uses credits~\cite{Cho2020Breakwater}. Svalinn and Protego operate inside one server~\cite{Pardeshi2026Svalinn,Cho2023Protego}. RPC stacks use CoDel and adaptive LIFO~\cite{Maurer2015FailAtScale}. In \system, refused work waits durably. Closest to \ride are concurrency limiters adapting to latency or rejections~\cite{NetflixConcurrencyLimits,EnvoyAdaptiveConcurrency} and clients throttling proportionally to rejections~\citep[``Client-Side Throttling'']{Cuervo2016HandlingOverload}. But the limiters read rising latency or drops as overload. An adaptive limiter could set an inference engine's cap (untested). Latency-driven dispatch needs latency to rise with load, but a fail-fast \bound refuses excess immediately, flattening that signal. \ride, one MIMD~\cite{Chiu1989AIMD} controller per capacity domain, separates refusals from timeouts and cuts deeper as more attempts are refused, as DCTCP~\cite{Alizadeh2010DCTCP} does with marks. Other schemes signal congestion by delay, drops, or queue sojourn time~\cite{Brakmo1994TCPVegas,Floyd1993RED,Nichols2012CoDel}. CONCUR sets agentic batch-inference admission from KV-cache signals inside the engine, with a controller inspired by congestion control~\cite{Chen2026CONCUR}. A client-side scheduler cuts HTTP 429s against shared API quotas, without targeting LLM serving~\cite{Farkiani2026ClientRateLimit}.

\smallskip\noindent\textbf{ML and LLM serving.}
Autoscalers~\cite{KEDA,Fu2024ServerlessLLM,Zhang2025BlitzScale,Jaiswal2025SageServe,Patke2025Chiron} add capacity in steps. Engine schedulers and batch runtimes~\cite{Gujarati2020Clockwork,Yu2022Orca,Kwon2023PagedAttention,Agrawal2024SarathiServe,Zhong2024DistServe,Patel2024Splitwise,Sun2024Llumnix,Zhang2026JITServe,Xu2026BatchGen,Goel2026QoServe} change what runs in or across engines. \system sets how much work enters, complementing both. In production, offline LLM inference harvests idle online capacity~\cite{Yang2026ACDC}, and one provider's serving layer enforces admission and quotas~\cite{Raichura2026Bedrock}. QLM orders queued work by SLO~\cite{Patke2024QLM}, and Mooncake rejects early on predicted load~\cite{Qin2025Mooncake}. We refuse at the engine. Google's AIRS queues LLM rating requests, retrying failures with exponential back-off. AIRS matches dispatch to RPC success rates on shared TPU pools~\cite{Jagnik2026AIRS}. \ride adapts only to labeled capacity refusals (429s), excluding timeouts and other errors.

\section{Limitations}

Outside replay and the A/Bs, the production results are observational (Section~\ref{sec:evaluation}). The real-engine \ride testbed used the classifier request shape (Appendix~\apRevfixed). We evaluated generative workloads in production. On the emulator, two \ride controllers sharing one capacity domain split its capacity nearly evenly (Appendix~\apCtrl). Production avoids such sharing. We have not applied \system to interactive or agent-session traffic. Future work includes prefix-aware queue ordering and queue-depth autoscaling to drain peaks faster. Another item is routing requests grouped by sequence length to engines tuned for that length. When the online lane's queue runs low, routing batch work there would keep the committed reservation full.

\section{Conclusion}
\label{sec:conclusion}

\system paces dispatch as congestion control paces a network sender. A \bound refuses work early where capacity saturates. \ride probes upward and backs off on those refusals, as a sender does on loss. Network loss wastes work, but a refusal is lightweight feedback because the work waits durably and refused attempts use no reserved tokens. We report one implementation, Meta's traffic shaper, carrying more than 300 million online requests a day. Same-traffic A/Bs compared the shaper (caps plus \ride) with its precursors. The shaper served a classifier on 43\% fewer H100s and, on an equal fleet, drained zero-shot backlogs a median 3 hours sooner. In steady-capacity replay, \ride came within \rqAbsGap of the goodput of a fixed rate tuned above capacity, with \rqRejRatio of that rate's refusal share. After the provider's admission update, the shaper averaged 95.8\% of reserved provider capacity over 15 weeks. A backlogged week averaged 99.1\%. On the classifier's smaller fleet, lower-priority work waited longer at backfill peaks. \ride trades a slower climb after deep drops for smaller swings at a stable ceiling.

Practitioners should bound the inference engine's queue, size the fleet for freshness, and make the simple path the only path. In a layered serving stack, a framework acts as the \bound if it caps each replica's in-flight requests~\cite{RayServeMaxOngoing,KnativeConcurrency} and refuses beyond a bounded queue~\cite{RayServeMaxOngoing,TritonQueuePolicy}. A cluster scheduler with quotas and preemption allocates each workload GPUs for its typical load and lends a shared flex pool to backlogged workloads. Queue depth behind \ride shows where the pool is needed, as in event-driven autoscalers~\cite{KEDA}. We have not evaluated \system on this stack. The approach suits latency-tolerant work behind backends that refuse promptly.

\bibliography{references}
\bibliographystyle{mlsys2025}

\section*{AI Usage Disclosure}

We conceived \system to address the capacity limits, spiky traffic, and over- and underutilization we saw in production. We designed its architecture. We used LLM-based assistants (Claude, GPT, and an internal Meta model) through Meta's internal developer tools, as follows.

\emph{Design and code.} We asked AI assistants to review our architecture and find weaknesses in it. Engineers and AI coding assistants then wrote the system code together, and engineers reviewed it through normal human code review, assisted by AI review tools.

\emph{Experiments.} We designed the controlled experiments, including their arms, phases, metrics, and preregistered analyses. AI agents set up the test harness around our production code on our GPU machines, executed the runs, gathered the logs, and cross-checked the logs with the results. AI agents also wrote the emulator and simulator around our production code and generated the synthetic replay prompts to our specifications. We reviewed the test harness code and every reported result, and relied on the agents' cross-checks for the gigabytes of raw logs.

\emph{Data and analysis.} We chose what production data to gather. AI assistants wrote and ran some of the queries against our internal systems and computed the reported statistics. We also gathered data across our systems by hand. We checked all the outputs.

\emph{Literature.} AI assistants helped search for and map related work. We retrieved each cited work, took its bibliographic entry from the publisher or project page, and checked that it supports the sentence that cites it. An AI-assisted audit then rechecked every citation.

\emph{Writing and figures.} We wrote the outlines and section drafts by hand. AI assistants helped review, reorder, and reframe each section, check accuracy and consistency, and adjust wording toward our goals. We then reviewed and edited the text, going back and forth with the assistants over many rounds. AI assistants also produced the \LaTeX{} formatting and drew the figures from our data.

Every reported number comes from our production measurements and experiment logs.

\clearpage
\appendix
\renewcommand{\thetable}{\Alph{section}\arabic{table}}
\makeatletter\@addtoreset{table}{section}\makeatother
\renewcommand{\thefigure}{\Alph{section}\arabic{figure}}
\makeatletter\@addtoreset{figure}{section}\makeatother

\section{Production Details}
\label{app:production}

\subsection{Batch lane and scale}
\label{app:batch}\label{app:scale}

The batch lane serves multiple applications and is counted separately from all online volumes in this paper. Over 20 days before the load test, it submitted more than 50 million requests a day. Two workloads joined during the one-day production load test. They were a second fine-tuned model on MI350X GPUs and another model on a fourth accelerator type. No controller setting changed that week, even though one workload's owner had cut output and reasoning tokens per request nearly five-fold four days earlier. On the test day, permanent errors stayed at or below 0.06\% of completions on every workload. The week after the load test, the four accelerator fleets' hourly peak allocations summed to more than 4,000 accelerators.

Three weeks before the load test, online volume was about 160 million requests a day, excluding the batch lane. During a later backfill on the MI350X workload, the shaper carried 1.125 billion online requests in its busiest 24 hours. Over the five complete days after the backfill drained, it carried 308--351 million online requests a day. New workloads keep joining, some beyond the social media applications.

A workload rode its ceiling in an hour when its queue averaged at least 10,000 items and its oldest item had waited at least 10 minutes. At least 0.5\% of attempts also had to be refused. Over the classifier backfill's middle two days, the classifier and zero-shot workloads, each behind per-replica caps, rode their ceilings for all 48 hours. The reserved tier, behind the provider's admission, rode its ceiling for 34 hours. We varied the queue-size threshold from 1,000 to 1 million items and the oldest-item-age threshold from 1 to 60 minutes. The refusal-share threshold ranged from 0.1 to 2\%. Under all 36 combinations, the classifier and zero-shot workloads rode 47--48 of the 48 hours and the reserved tier 29--35. In those two days, the median hourly refusal share while each workload rode its ceiling was 7.8\% for the classifier, 6.4\% for zero-shot, and 4.1\% for the reserved tier.

\subsection{Classifier and zero-shot A/Bs}
\label{app:fleetsrc}

\emph{Classifier.} The shaper and its precursor each scored every classifier request on full production traffic, with the same model, prompts, and vLLM settings. Profiling sized the precursor at 68 H100s. The precursor's inference engines began to error just above its usual online load. That error point was about two-thirds of the profiled per-card rate, so with no margin the precursor would have needed 63 H100s. The shaper carried the same traffic on 39 H100s. At that load, each shaper card ran slightly above the profiled per-card rate. Both paths drained \foqs-backed queues. Of the 29 fewer H100s, 5 are the margin that the precursor's 68-card fleet carried. The other 24 come from the shaper's engines running at their plateau behind the caps and \ride. The launch review found 100\% coverage parity and identical scores up to LLM sampling noise. This A/B tests the whole design. Replay (main text, Sections 6.1 and 6.2) isolates each part.

We compare hourly end-to-end latency percentiles for each priority band on each path, over matched hours on eight days spanning the classifier backfill. Outside the backfill's two peak days, the ratio of shaper to precursor latency was at most 1.04 on each compared day, in every band at p50, p95, and p99. On the median day, the ratio was 0.997--1.006. Over the two peak days, the shaper's top-band p99 in the classifier stage was 0.35 times the precursor's (ratio of 48-hour means). But the shaper's lowest band waited a median of about 6.6 hours in the classifier stage, against the precursor's half hour. Each shaper card served about 1.6 times the rate at which precursor cards began to error. At equal size, the shaper would then drain a backlog about 1.6 times as fast (an untested estimate). A pool that grows with queue depth could also absorb that wait. After a resize, the shaper's fleet ran 45 H100s over the last 15 days of the three-month arrival window, against an average need of 30.

An overload episode hit the classifier before its caps were enabled, under a 600~s timeout (now 100~s). In the episode's worst four minutes, 10.2\% of requests timed out (sample-weighted).

\emph{Zero-shot.} Over the 23-day A/B, both paths used the same priority ordering on equal MI300X fleets. Daily averages were 601.7--610.1 GPUs for the precursor and 595.1--609.3 for the shaper. With leased items included, both paths were backlogged at eight peaks. The shaper drained first at all eight, by a median of 3 hours.

Latency comes from a 5\% sample of the same posts on both paths. Peak hours are the 135 hours when either queue's oldest item was over an hour old. In those hours, the shaper's p99 in the top three bands was 3.4--3.6 minutes, against the precursor's 6--8 minutes. In the lowest band, it was about an hour against about 8 hours. The shaper's median in the top three bands was about 8~s against about 2 minutes, and in the lowest band 8~s against about an hour. Off-peak hours are the 399 hours when both queues' oldest items were under 10 minutes old. In those hours, the shaper's median was 5.5~s against 2.1~s. Platform steps before the shaper's dispatcher account for about 2.4~s of the shaper's median.

\subsection{Fleet sizing and autoscaling}
\label{app:autoscale}

Hourly arrivals come from queue conservation, which can only undercount. A serve-on-arrival fleet serves each hour's arrivals with no headroom. Its replicas run at the per-replica rate that engines held behind the caps, so its size is a lower bound. Over three months, serving mean classifier arrivals this way needed 25 H100s. The busiest hour reached 4.0 times the mean, or 99 H100s. Autoscaling narrows the gap between mean and peak only as fast as it reacts. To keep up with the sharpest one-hour rise, an unqueued fleet that reacts within an hour would need at least 74 more H100s than a serve-on-arrival fleet sized for the preceding hour. With the queue, rises in arrivals became backlog.

Over three months, the internal fleets changed size nine times. The classifier fleet changed five times (12 to 34, 34 to 44, 44 to 34, 34 to 39, and 39 to 45 H100s), and the zero-shot fleet four times. A new replica takes a median of 136~s (classifier) to 159~s (zero-shot) to load its model and serve. The classifier grew from 34 to 39 and 39 to 45 H100s using quota already held. The added capacity was serving 27 and 13 minutes after each change landed, excluding review.

With the zero-shot caps refusing work, throughput per replica-hour was 4\% lower at 300--305 replicas than at 153--155. The larger fleet ran on the A/B's second day, the smaller 23 days earlier. Admission followed the zero-shot fleet as it grew from about 154 replicas to 302. Over days 2--17 of the A/B, the elastic pool preempted the zero-shot tenant's replicas 4,163 times, acting like an autoscaler that takes capacity back. In that period, 0.0004\% of the tenant's requests timed out.

Figure~4 of the main text shows the three mornings of on-demand losses. Each morning, completions fell from 90--170\% of their pre-loss median to 1.6--3.3\%. During each loss, \ride's target rate sat at its floor, and the queue grew to 8--11.5 hours of median completions. Completions then peaked at up to 12 times the median. No capacity setting changed. When the pool's own capacity losses left results arriving too late, the workload moved to a higher tier that refuses less, where \ride follows the provisioned capacity.

\subsection{Per-replica caps and global limits}
\label{app:percap}

The cap admits 12 requests beyond vLLM's \texttt{max\_num\_seqs}. These extra requests keep each inference engine's batch full across a dispatch round trip. At the measured per-replica throughput, they wait about 8~s on a zero-shot replica, against a 600~s timeout. On a classifier replica they wait under half a second, against a 100~s timeout. The change that set the caps also raised vLLM's GPU memory allocation on the shaper's zero-shot engines from 80\% to 85\%. At the engines' KV cache use (below), the allocation change made no difference. On the classifier backfill's middle two days, 97.5\% of live replicas refused excess work in the median hour.

Each precursor dispatched its job type under one global rate limit, retuned by hand six times in about four months. The zero-shot precursor's migration review found that a high limit caused tenant timeouts and retry storms, while a low limit left GPUs idle.

Over days 2--17 of the zero-shot A/B, both paths ran at their dispatch limits in 81 hours. In each, the precursor's limit throttled dispatch and the shaper's caps refused at least 0.1\% of requests. In those hours the shaper's engines ran a median of 123.8 of their 128 sequences, against the precursor's 40.7. KV cache use was 31.6\% for the shaper against 13.2\% for the precursor. In every shaper region with at least nine engines, 89--100\% of engines had an hourly mean of at least 120 running sequences. Both zero-shot paths ran the same model on every eligible post, with 17.3 and 17.4 input tokens per output token. The shaper served 1.09 times the precursor's model requests per GPU. At that ratio it could carry the precursor's daily load on about 48 fewer of about 600 GPUs, an untested estimate. While its limit was binding, the precursor ran at a median 86\% of the peak request rate both fleets reached.

\subsection{Priority and freshness}
\label{app:fresh}

The classifier's bands and their freshness goals are top (30 minutes), high (3 hours), middle (12 hours), and low (24 hours). Over one month, these bands completed 96.0, 94.1, 69.0, and 82.7\% of items within their goals. Over the backfill's last three days, the same shares were 99.3, 89.3, 44.2, and 55.1\%. The lower two bands, at 44--55\%, absorbed the wait. Over 12 days outside backfills, the oldest undelivered item in the classifier queue was a median 1~s old.

\subsection{Reserved capacity}
\label{app:backlog}\label{app:reserved}

The static system had 18 full days in the 23 days before launch, outside the shaper's earlier two-day test and pre-launch test. Its daily means of six-hour utilization ranged 71.04--94.38\%. The shaper's four full launch days ranged 93.67--96.28\%, and each was exceeded by at most one static day. We raised \ride's global target refusal share in steps from 5\% at launch to 30\% over that day and the next two. The usual token mix was 2.9 input tokens per output token in the three weeks before launch and 3.0 in the launch week. The static system's 88.2\% utilization left 11.8\% of the reservation unused. The shaper's 94.8\% left 5.2\% unused. The shaper's daily refusals were 23--33\% over the 20 days after launch. Its refusal share over the first four days was 28.3\%. Their daily mean over the 20 days was 27.6\%. The static system's overall refusal share was 16.1\% over the 23 days before launch, outside test days.

Over three days, 15-minute intervals with more than 1,000 items in the reserved workload's queue averaged 98.6\% utilization. Intervals with 1,000 or fewer averaged 88.6\%. Over the following three weeks, all 84 hours with fewer than 1,000 queued items fell in the evening and at night. Over the eight days from the day before the load test, the backlog was continuous, and utilization stayed above 97\% in every six-hour period (98.9\% mean). In the backlogged week, at least 114K items were queued in every hour after the first two, and utilization averaged 99.1\%. Over the 15 weeks after the provider's update, request arrivals swung daily by about 2.5 times between the quietest and busiest tenth of hours. Completions varied by under a sixth because the queue absorbed the swing. Without the queue, peak-sized capacity would sit idle off peak.

The static system's 88.2\% mean utilization uses six-hour samples over the 23 days before launch. Its median hourly refusal share was 24.4\% over its busiest quarter of hours in the last eight days before launch. The shaper's was 4.4\% over the 15 weeks after the provider's update. Twenty days after launch, the provider's admission rose from near 95--97\% to the full reservation. The shaper then used a 2\% target refusal share and 3\% rate increases.

\subsection{The pre-launch test}
\label{app:dialup}

At a low target refusal share, backing off on refusals sends the provider far fewer refused attempts than a fixed rate does. Within two hours of the shaper taking over, the refusal share across both dispatchers fell from 22.6\% to 2.8--4.9\%. While the shaper ran alone, an unintended upstream change to media resolution shifted the mix to 1.75 input tokens per output token. The static system's mix had been 2.89 in the six days before. After a 12-hour ramp under that output-heavy mix, the shaper held 97--99\% of the reservation. Refusals were near its 5\% target refusal share. After the static system resumed, the refusal share rose to 14.0--21.0\% in two-hour bins that evening. In the overlaps (10 hours at the start and 8 at the end), the two dispatchers together completed about as many requests per hour as either alone. Input and output token rates returned within 1\% of their pre-test levels in the six days after the test.

About 7\% of test requests queued and timed out in our gateway's smallest region before reaching the provider. The gateway routed only within a region, while the shaper spread work evenly across three regions. Waiting without a refusal hid the overload until requests timed out, as in ``Queues inside the server'' (main text, Section 3).

\subsection{Four kinds of \bound}
\label{app:bounds}\label{app:refusal}

\ride ran behind four kinds of \bound in production. \emph{Model-server caps} on the internal fleets refuse work once a replica's in-flight count reaches its cap. The model server then raises an overload exception, and the RPC layer marks the response as an overload. The gateway maps only overload statuses to HTTP 429. \emph{Provider admission} on the reserved tier is opaque to us. The provider's admitted ceiling moved with the token mix. Under the usual mix, utilization averaged 93--95\% of the reservation at refusal shares up to 33\%. Under the pre-launch test's output-heavy mix, utilization reached 97--99\% (Appendix~\ref{app:dialup}).

During the load test, a reactive \emph{per-caller token quota} in our gateway limited the model on the fourth accelerator type. In minutes when the model drew at least 90\% of its quota, a median 14\% of attempts were refused. Over the week of the load test, the shaper's refusals tracked the gateway's blocks, with a per-minute correlation of 0.97. The gateway's audit log attributed all but 0.01\% of the model's recorded refusals to the quota. In the load test's six hours with refusals, the router in front of the model's inference engines recorded zero waiting requests at the engine for all but 0.001\% of requests. The accelerators were busy 78--89\% of the time. Median latency held near 11~s, with hourly medians of 10.4--12.5~s. Attempts timed out in a single minute of the week.

An \emph{engine queue-depth aware router} in front of one of our production tiers balances requests across inference engines by load. The router refuses with HTTP 429 once an engine's waiting queue passes a limit. In a staging stress test, \ride dispatched to 24 engines at about four times the load those engines could sustain. With a limit of 300 waiting requests, every engine's waiting queue held near 300, and 99.9\% of requests succeeded. We then raised each engine from 256 to 320 sequences and 8,192 batched tokens. The larger batch gained 10--15\% throughput. But with a deep waiting queue, the larger batch timed out too often to deploy. At 320 sequences, we lowered the waiting limit from 300 to 100. The same 19,200 requests then finished in 1,041~s instead of 1,400~s. The lower limit also cut timed-out attempts from 15.4\% to 1.6\%. Requests that saw any timeout fell from 23.8\% to 1.8\%.

\subsection{Internal overload episodes}
\label{app:episodes}

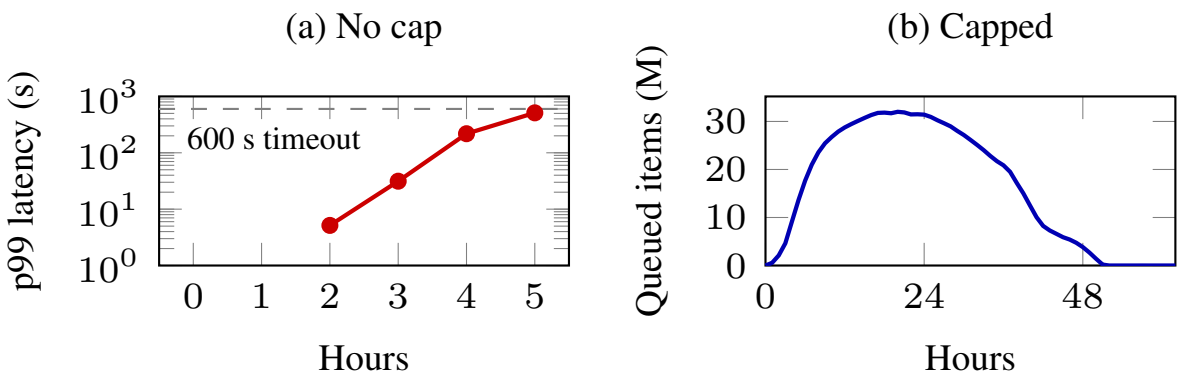
\begin{figure}[t]
\centering
\begin{tikzpicture}
\begin{axis}[name=nocap, width=0.52\columnwidth, height=2.7cm, xlabel={Hours}, ylabel={p99 latency (s)},
  ymode=log, ymin=1, ymax=1000, xmin=-0.5, xmax=5.5, xtick={0,1,2,3,4,5},
  tick label style={font=\scriptsize}, label style={font=\scriptsize}, title style={font=\scriptsize}, title={(a) No cap}]
\addplot[mark=*, mark size=1.2pt, thick, red!80!black] coordinates {(2,5.15) (3,31.34) (4,218.04) (5,511.39)};
\addplot[dashed, gray, domain=-0.5:5.5] {600} node[pos=0.28, below, font=\tiny, black] {600 s timeout};
\end{axis}
\begin{axis}[at={(nocap.east)}, anchor=west, xshift=1.3cm, width=0.52\columnwidth, height=2.7cm,
  xlabel={Hours}, ylabel={Queued items (M)}, ymin=0, xmin=0, xmax=62,
  xtick={0,24,48}, tick label style={font=\scriptsize}, label style={font=\scriptsize}, title style={font=\scriptsize}, title={(b) Capped}]
\addplot[thick, blue!70!black] coordinates {(0,0.006) (1,0.558) (2,2.099) (3,4.623) (4,9.165) (5,13.661) (6,17.687) (7,20.996) (8,23.566) (9,25.448) (10,26.750) (11,27.885) (12,28.762) (13,29.467) (14,30.117) (15,30.744) (16,31.337) (17,31.776) (18,31.837) (19,31.707) (20,32.008) (21,31.878) (22,31.464) (23,31.495) (24,31.418) (25,30.945) (26,30.253) (27,29.629) (28,28.975) (29,28.038) (30,27.175) (31,26.163) (32,25.121) (33,24.027) (34,22.824) (35,21.734) (36,20.863) (37,19.508) (38,17.208) (39,15.034) (40,12.522) (41,10.058) (42,8.242) (43,7.297) (44,6.600) (45,5.899) (46,5.415) (47,4.738) (48,3.834) (49,2.694) (50,1.405) (51,0.191) (52,0.000) (53,0.000) (54,0.000) (55,0.000) (56,0.000) (57,0.004) (58,0.000) (59,0.000) (60,0.000) (61,0.000) (62,0.000)};
\end{axis}
\end{tikzpicture}
\caption{Internal classifier path. (a) Without the cap, hourly-averaged p99 latency climbed toward the 600-second timeout then in force. (b) With the cap, the four-day backfill waited in the queue (peak 32M items, x-axis in hours from its first queued items) while nearly every replica refused excess work. The panels carry different loads.}
\label{fig:firstparty}
\end{figure}

Before the model-server caps were set, bursts of classifier timeouts recurred 5 and 10 days after the first burst (Figure~\ref{fig:firstparty}). In the first burst, two replicas timed out every request, one for four minutes and the other for two. On the day of the second recurrence, an edit set the request and processing caps to 140. The first overload refusal came two days later. None of the following 151 hours recorded a timeout. The zero-shot tenant followed the same sequence about six weeks later. On two consecutive days, timeouts hit 10.8\% and 16.3\% of its requests. The first refusals came on the second day. Under 0.01\% of requests timed out on each of the next 16 days. Later, with the caps in place, a four-day classifier backfill queued up to 32M items. The backlog drained within 32 hours of the peak. On the backfill's middle two days, at least 81\% of the roughly 40 live replicas refused excess work in every hour.

\section{Coordination}
\label{app:coordination}

\begin{figure}[t]
\centering
\begin{tikzpicture}[x=1cm, y=1cm, >=stealth, font=\scriptsize,
  q/.style={draw, fill=white, rounded corners=4pt, align=center, inner sep=2.5pt},
  act/.style={draw, fill=red!12, rounded corners=1pt, align=center, inner sep=2.5pt, minimum width=1.05cm},
  hold/.style={draw, fill=gray!10, rounded corners=1pt, align=center, inner sep=2.5pt},
  yn/.style={font=\tiny, inner sep=1.5pt},
  flow/.style={->, thick}]
\node[q, fill=green!14] (start) at (3.2,0) {Every 15 s, for each queue: read fleet-wide\\successes, 429s, and failures since the last poll};
\node[q] (d1) at (2.4,-0.95) {New outcomes this poll?};
\node[hold] (h1) at (6.3,-0.95) {Hold};
\node[q] (d2) at (2.4,-1.8) {Worker count changed\\in the last 10 polls?};
\node[hold] (h2) at (6.3,-1.8) {Freeze (hold)};
\node[q] (d3) at (2.4,-2.85) {429s $\le$ 2\% of\\successes + 429s?};
\node[q] (d4) at (1.0,-3.75) {Dispatch $\ge$ 75\%\\of target?};
\node[q] (d5) at (5.3,-3.75) {429 share};
\node[act, fill=green!25] (inc) at (0.4,-4.85) {+3\%\\($\ge$0.1/s)};
\node[hold] (gate) at (1.75,-4.85) {Hold (gate)};
\node[act] (m1) at (4.0,-4.85) {$\le$50\%:\\$\times$0.99};
\node[act] (m2) at (5.3,-4.85) {50--65\%:\\$\times$0.85};
\node[act] (m3) at (6.6,-4.85) {$>$65\%:\\$\times$0.5};
\draw[flow] (start.south -| d1) -- (d1);
\draw[flow] (d1) -- node[yn, above] {no} (h1);
\draw[flow] (d1) -- node[yn, right, pos=0.4] {yes} (d2);
\draw[flow] (d2) -- node[yn, above] {yes} (h2);
\draw[flow] (d2) -- node[yn, right, pos=0.4] {no} (d3);
\draw[flow] (d3.west) -| node[yn, left, pos=0.75] {yes} (d4.north);
\draw[flow] (d3.east) -| node[yn, right, pos=0.75] {no} (d5.north);
\draw[thick] (d4.south) -- (1.0,-4.3);
\draw[flow] (1.0,-4.3) -| (inc.north);
\draw[flow] (1.0,-4.3) -| (gate.north);
\node[yn, anchor=east] at (0.33,-4.45) {yes};
\node[yn, anchor=west] at (1.82,-4.45) {no};
\draw[thick] (d5.south) -- (5.3,-4.25);
\draw[flow] (5.3,-4.25) -| (m1.north);
\draw[flow] (5.3,-4.25) -- (m2.north);
\draw[flow] (5.3,-4.25) -| (m3.north);
\end{tikzpicture}
\caption{\ride's per-queue decision, made by the leader every 15 seconds, with the settings during our measurements.}
\label{fig:aimd}
\end{figure}
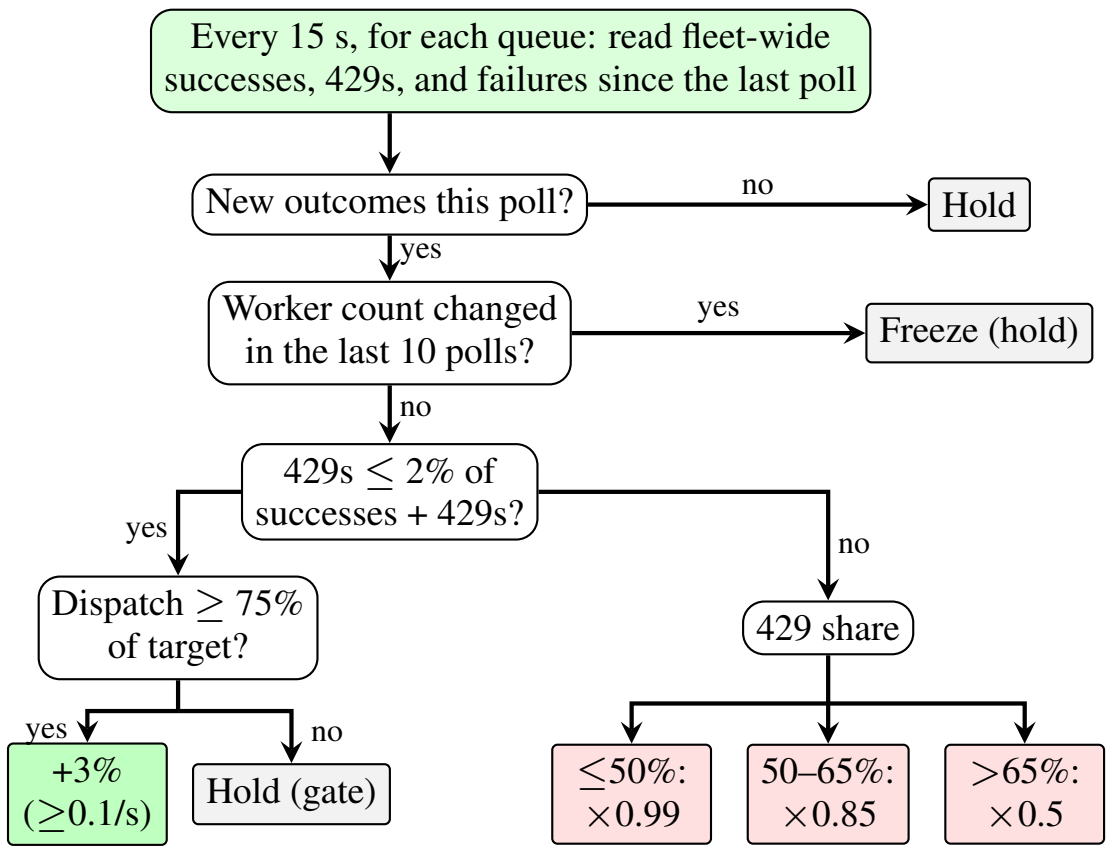

\subsection{Mechanisms}
\label{app:mechanisms}

\emph{Leadership.} The traffic shaper runs 45 workers in seven clusters. Every 30 seconds each worker asks fleet discovery for the current workers. A worker that cannot reach fleet discovery keeps its last-known role. The hash ring is deterministic, so a new leader computes the same placement.

\emph{Placement.} Each queue gets $\max(3, \min(N, \lceil r/q \rceil))$ of the $N$ workers, where $r$ is the queue's rate and $q$ a configured per-worker rate.

\emph{Draining.} A draining worker stops fetching and finishes what it holds. It declines the leader's polls, so placement rebalances at once.

\emph{Freeze.} When the number of responding workers changes, the leader still moves assignments but holds rates until 10 polls (2.5 minutes) after the last change (Figure~\ref{fig:aimd}). Section 5 of the main text describes the hysteresis. With it, only a drop of at least 10\% below the largest worker count in the last 20 polls starts a freeze. A smaller change, such as one worker missing a poll, holds rates for that poll only.

\emph{Delivery.} Interrupted work returns through \foqs lease redelivery after 20 minutes. Within a lease, the worker retries a request after a 429 or a timeout. Requests with other retryable errors are tried at most three times. The worker persists each response before acknowledging its input, so delivery is at least once unless storing a response fails.

\subsection{Safeguards in production}
\label{app:safe}

In one month before the hysteresis, all 2,205 worker-count changes triggered the freeze. No rate changed during the 10,011 frozen polls. With 15-second polls, rates froze for about 42 hours. Rolling deploys proceed in steps, and the leader can adjust rates between freezes. Capacity changes wait until the freeze lifts. During the wait, caps and provider 429s refuse excess work, preserving safety. The wait risks efficiency. We grouped consecutive target-rate decreases within 60 seconds into bursts. In one logged week, only 14 bursts across four queues cut the target by 30\% or more. We deploy a 3\% increase step because such deep cuts were rare.

\section{Controlled Experiments}
\label{app:controlled}

\begin{table*}[t]
\caption{\ride versus fixed rates on real engines, median of \rqRuns runs per cell ($^{*}$rate fixed at cut capacity, cap 40 only). Goodput is requests per second per host. Refusals are refused attempts per completion. Recovery is time from restore to 95\% of full capacity. Latency is the whole-run median from first send to success in seconds. A dash marks an arm that never recovered.}
\label{tab:revfixed}
\vspace{4pt}
\centering
\scriptsize
\setlength{\tabcolsep}{4pt}
\begin{tabular}{llrrrrrrrr}
\toprule
 & & \multicolumn{2}{c}{Full capacity} & \multicolumn{2}{c}{Cut} & \multicolumn{2}{c}{Restored} & Recovery & Latency \\
Cap & Arm & goodput & refusals & goodput & refusals & goodput & refusals & (s) & (s) \\
\midrule
40 & \ride & \rqTARideFullG & \rqTARideFullR & \rqTARideCutG & \rqTARideCutR & \rqTARideRestG & \rqTARideRestR & \rqTARideRec & \rqTARideQue \\
40 & Fixed 100\% & \rqTAFixFullG & \rqTAFixFullR & \rqTAFixCutG & \rqTAFixCutR & \rqTAFixRestG & \rqTAFixRestR & \rqTAFixRec & \rqTAFixQue \\
40 & Fixed 120\% & \rqTAHighFullG & \rqTAHighFullR & \rqTAHighCutG & \rqTAHighCutR & \rqTAHighRestG & \rqTAHighRestR & \rqTAHighRec & \rqTAHighQue \\
40 & Fixed at cut$^{*}$ & \rqTALowFullG & \rqTALowFullR & \rqTALowCutG & \rqTALowCutR & \rqTALowRestG & \rqTALowRestR & \rqTALowRec & \rqTALowQue \\
\midrule
64 & \ride & \rqTBRideFullG & \rqTBRideFullR & \rqTBRideCutG & \rqTBRideCutR & \rqTBRideRestG & \rqTBRideRestR & \rqTBRideRec & \rqTBRideQue \\
64 & Fixed 100\% & \rqTBFixFullG & \rqTBFixFullR & \rqTBFixCutG & \rqTBFixCutR & \rqTBFixRestG & \rqTBFixRestR & \rqTBFixRec & \rqTBFixQue \\
64 & Fixed 120\% & \rqTBHighFullG & \rqTBHighFullR & \rqTBHighCutG & \rqTBHighCutR & \rqTBHighRestG & \rqTBHighRestR & \rqTBHighRec & \rqTBHighQue \\
\bottomrule
\end{tabular}
\end{table*}

\subsection{Replay setup}
\label{app:replay}

The replays in Appendices~\ref{app:codel} and~\ref{app:budget} run Llama-3.1-8B-Instruct with vLLM 0.23.0 on one A100 80 GB, behind production builds of the model server and gateway. The uncapped arm disables the model server's request cap. The model server's semaphore instead blocks requests beyond the capped arm's count, so the arms differ only in blocking versus refusing. The load generator runs the production worker's dispatch loop at a fixed rate with Poisson arrivals and fixed seeds. It retries timed-out attempts as in production. Each arm ran three times on one trace.

\emph{Classifier.} Goodput plateaus at 9.14 requests per second and reaches 95\% of that plateau at 8 concurrent requests. Offered load is 17.5 attempts per second, and the timeout is 100~s.

\emph{Generative.} Goodput plateaus at 1.85 requests per second and stays within 3\% of that plateau from 32 to 192 concurrent requests. Offered load is 3.68 attempts per second (5.52 in the token budget runs), and the timeout is 600~s.

\subsection{No cap, deadline drop, and CoDel}
\label{app:codel}

Without the cap, steady-state goodput fell to zero from about minute 4 for the classifier and minute 20 for the generative shape (main text, Figure~2). Requests waited in the semaphore until they outlived the timeout. For the classifier, the median wait was 172--180~s against a 100~s timeout. For the generative shape, the median was 660~s against 600~s. The inference engine ran at its plateau in both arms (14,557 classifier tokens per second capped against 14,572 uncapped). With the cap, goodput held at 8.64 (classifier) and 1.83 (generative) requests per second. The excess waited in the dispatcher's queue instead, for a median of 177~s (classifier) and 744~s (generative).

Without the cap, we also dropped each request as it left the semaphore if its wait had already reached the caller's timeout. Steady-state goodput stayed at 0.01 requests per second for the classifier and 0 for the generative shape. Canceling calls still queued at 1.1 times the timeout, before they reached the engine, left classifier goodput at 0. The generative uncapped arm cancels this way throughout.

A CoDel-style rule used a target delay of about one median service time, 830~ms for the classifier and 17.2~s for the generative shape. The rule watched the oldest queued request's wait. Once that wait had stayed at or above the target for ten target delays, the rule refused new requests with the cap's overload status until the queue drained. The rule matched the cap's goodput with no client timeouts (classifier 0.3\% above the cap, generative 1.82 against 1.83).

\subsection{Cap level, token budget, and vLLM queue limits}
\label{app:budget}

We raised the cap to twice and four times each shape's plateau concurrency, which is 8 requests for the classifier and 32 for the generative shape. Median goodput changed by +2.0\% and +2.4\% for the classifier, and by +0.5\% and +0.4\% for the generative shape. The excess queued inside the engine, and service time rose two- to fourfold. In place of the cap, a per-replica token budget equal to eight mean-sized classifier requests matched the cap's classifier goodput. That token budget also lowered p99 service latency by 27\%. On the generative shape, neither of two token budget sizes gave a clear gain over the 32-request cap. vLLM 0.29 added \texttt{-{}-max-num-queued-reqs} and \texttt{-{}-max-num-queued-tokens}, which are off by default and refuse with HTTP 503.

\subsection{\ride versus other controllers on an emulated provider}
\label{app:ctrl}

Four queues shared a five-worker test fleet running the deployed controller logic and settings. A mock provider capped each queue's requests per second over a 10-second window. It refused excess attempts with a 429 at once and served the rest after a per-queue service time that does not grow with load. In each 40-minute run, one queue's capacity dropped from 20 to 5 requests per second at minute 10 and returned at minute 20. Across arms, that queue used \ride, a rate fixed at the pre-drop capacity, or classic additive-increase/multiplicative-decrease (AIMD) control (halving the rate on any poll above the threshold). Other arms used AIMD with \ride's decrease and an increase of 1 request per second per step, \ride with a 10\% step, or Gradient2 adapted to limit the rate by charging each refused attempt one service time. Each arm ran three times.

During the drop, \ride drew 0.25 refused attempts per completion. AIMD with \ride's decrease drew 0.30, the 10\% step drew 0.29, and classic AIMD drew 0.12. The fixed rate drew 2.87 and Gradient2 drew 3.63. Before the drop, classic AIMD served about 66\% of capacity, and the other three low-refusal arms served about 96\%. Gradient2's only latency signal was the service time charged to each refused attempt. Its target rate reached about 2.5 times capacity.

The step size trades recovery speed against swing. Per-queue rates in production run from a few requests per second to about 16,000. A step of 1 request per second per poll would take two hours to double a rate from 500 to 1,000 requests per second. A step sized for fast queues overshoots slow ones. A 3\% step doubles any rate within about 24 polls, or six minutes. The 10\% step and AIMD with \ride's decrease brought their target rates to 95\% of capacity in 217~s, against 671~s for \ride's 3\%. A discrete-event simulation ran the deployed controller source against the emulator's admission window and service times and reproduced these recovery times (213 and 677~s). With capacities scaled 25 times, the 10\% step's target rate swung 0.222 of capacity at a stable ceiling, against 0.065 for \ride.

In three separate runs, two queues, each under its own \ride, shared one capacity domain capped at 20, then 5, then 20 requests per second. The two controllers split the domain nearly evenly. Median Jain's index on completions was at least 0.998 in every phase.

\subsection{\ride versus fixed rates on real engines}
\label{app:revfixed}

Three identical hosts each ran two replicas of Llama-3.1-8B-Instruct, one per A100 80 GB. The replicas ran on production builds of the model server and vLLM (\texttt{max\_num\_seqs} 16, 16,384 batched tokens). Prompts were synthetic, with the classifier trace's request sizes shortened to a mean of about 247 tokens and one-token outputs. At these sizes, each replica served about 54 requests per second. Each 40-minute run began with a 5-minute warm-up and a full-capacity phase. A 10-minute cut then removed one replica from dispatch until the restore. The request timeout was 100~s. Calibration put each host's full capacity at about 108 requests per second and cut capacity at about 54.5. The model server held about 16 of each replica's in-flight requests, and vLLM held the rest. We tested caps of 40 and 64. At a cap of 40 a small queue formed in vLLM behind a full batch. At 64 about two batches waited. Table~\ref{tab:revfixed} reports medians of \rqRuns runs per cell. In the full-capacity phase, median request latency was 0.53~s under \ride against 0.64~s under the fixed 120\% rate. \ride's latency falls between those of the fixed 100\% and 120\% rates because it probes just above the ceiling. Under \ride, a few requests wait in vLLM, and refused attempts wait to retry.

\subsection{Ramping into an uncapped backend}
\label{app:ramp}

We ramped load into one classifier-shape replica, with and without the \bound. The replica's measured capacity, $X^*$, was 8.74 requests per second. The three increase rules were \ride's 3\%, $+0.05X^*$, and 10\%. We ran each rule three times. Each 40-minute run started at $0.5X^*$ with a 100~s client timeout. On the uncapped replica, each rule counted client timeouts as refusals. Behind the \bound, the rules reached 92.7--96.8\% of $X^*$. Uncapped, overload became visible only when timeouts arrived, a median 342--409~s after the server queue began to grow. Each rule's rate then dropped to its floor, and goodput fell to about zero while the queue drained. These stalls left each uncapped rule with 27.8--38.0\% less goodput than it had behind the \bound.
\end{document}